%% file: __main.tex
\pdfoutput=1
\documentclass[11pt]{article}

\input{acl/_default_preamble}
\input{_common_custom_preamble}
\input{acl/_custom_preamble}
\input{acl/_meta_info}

\begin{document}
\input{acl/top} 

\input{sections/1_intro}
\input{sections/2_background}

\input{sections/3_math_decomposition}

\input{sections/4_token_and_detok}

\input{sections/5_position_and_detok}

\input{sections/6_followup}

\input{sections/7_conclusions}
\input{sections/8_limitations}

\input{sections/9_ethics}

\input{sections/9_acknowledgements}


\bibliography{paperpile}

\onecolumn
\appendix
\input{sections/9_appendix}

\end{document}

%% file: acl/_default_preamble.tex
\usepackage[final]{acl}
\usepackage{times}
\usepackage{latexsym}
\usepackage[T1]{fontenc}
\usepackage[utf8]{inputenc}
\usepackage{microtype}
\usepackage{inconsolata}
\usepackage{graphicx}

%% file: _common_custom_preamble.tex
\definecolor{snsblue}{RGB}{59, 117, 175}
\definecolor{snsgreen}{RGB}{81, 158, 62}
\definecolor{snsorange}{RGB}{239, 134, 53}
\definecolor{snsred}{RGB}{234, 51, 35}

\definecolor{pptblue}{RGB}{89, 143, 163}
\definecolor{pptred}{RGB}{177, 86, 86}
\definecolor{pptgreen}{RGB}{116, 170, 156}
\definecolor{pptyellow}{RGB}{203, 163, 20}
\definecolor{pptdarkblue}{RGB}{67, 107, 122}

\usepackage{amsmath}
\usepackage{amssymb}
\usepackage{amsfonts}
\usepackage{bm}
\DeclareMathOperator{\diag}{diag}
\DeclareMathOperator{\var}{Var}

\usepackage{tcolorbox}
\tcbuselibrary{theorems,skins}

\newcommand{\redbox}[1]{
    \tcboxmath[
        colback=red!8,   
        colframe=blue!0,  
        top=-1pt,bottom=-1pt,left=-2pt,right=-2pt
        ]{ #1 }
}
\newcommand{\greenbox}[1]{
    \tcboxmath[
        colback=green!8,  
        colframe=blue!0, 
        top=-1pt,bottom=-1pt,left=-2pt,right=-2pt
        ]{ #1 }
}
\newcommand{\yellowbox}[1]{
    \tcboxmath[
        colback=yellow!10,   
        colframe=blue!0,   
        top=-1pt,bottom=-1pt,left=-2pt,right=-2pt
        ]{ #1 }
}

\newcommand{\blueboxannot}[2]{
    \tcboxmath[
        enhanced,         
        frame hidden,     
        colback=blue!5,
        top=-1pt,
        bottom=-1pt,
        left=-2pt,
        right=-2pt,
        overlay={
            \node[below,blue] at (frame.south) { #1 };
        }
    ]{ #2 }
}

\newcommand{\blueboxannottop}[2]{
    \tcboxmath[
        enhanced,       
        frame hidden,    
        colback=blue!5,
        top=-1pt,
        bottom=-1pt,
        left=-2pt,
        right=-2pt,
        overlay={
            \node[above,blue] at (frame.north) { #1 };
        }
    ]{ #2 }
}

\newcommand{\term}[2][]{$T^{\text{#2}}_{#1}$}

\usepackage{booktabs}

\newcommand{\linewidthcoef}{1.0}

%% file: acl/_custom_preamble.tex
\usepackage[capitalise,english,nameinlink]{cleveref}
\Crefname{fig}{\text{Figure}}{\text{Figures}}
\creflabelformat{equation}{#2\textup{#1}#3}
\crefname{line}{\text{line}}{\text{lines}}
\crefname{assumption}{\text{Assumption}}{\text{Assumptions}}
\usepackage{cite}

%% file: acl/_meta_info.tex
\title{Weight-based Analysis of Detokenization in Language Models:\\Understanding the First Stage of Inference Without Inference}

\author{
 \textbf{Go Kamoda\textsuperscript{1}}\hspace{1em}
 \textbf{Benjamin Heinzerling\textsuperscript{2, 1}}\hspace{1em}
 \textbf{Tatsuro Inaba\textsuperscript{3}}\hspace{1em}
 \textbf{Keito Kudo\textsuperscript{1}}
\\
 \textbf{Keisuke Sakaguchi\textsuperscript{1, 2}}\hspace{1em}
 \textbf{Kentaro Inui\textsuperscript{4, 1, 2}}
\\
\\
 \textsuperscript{1}Tohoku University\hspace{1em}
 \textsuperscript{2}RIKEN\hspace{1em}
 \textsuperscript{3}Kyoto University\hspace{1em}
 \textsuperscript{4}MBZUAI
\\
 \small{
   \textbf{Correspondence:} \href{mailto:go.kamoda@dc.tohoku.ac.jp}{go.kamoda@dc.tohoku.ac.jp}
 }
}

%% file: acl/top.tex
\maketitle
\begin{abstract}
\input{sections/_abstract}
\vspace{2.2ex}\\
\includegraphics[width=1em]{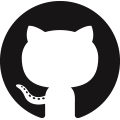}
\href{https://github.com/gokamoda/lm-detokenization}{github.com/gokamoda/lm-detokenization}
\vspace{0.5ex}
\end{abstract}

%% file: sections/_abstract.tex
According to the stages-of-inference hypothesis, early layers of language models map their subword-tokenized input, which does not necessarily correspond to a linguistically meaningful segmentation, to more meaningful representations that form the model's ``inner vocabulary''.
Prior analysis of this \emph{detokenization} stage has predominantly relied on probing and interventions such as path patching, which involve selecting particular inputs, choosing a subset of components that will be patched, and then observing changes in model behavior.
Here, we show that several important aspects of the detokenization stage can be understood purely by analyzing model weights, without performing any model inference steps.
Specifically, we introduce an analytical decomposition of first-layer attention in GPT-2.
Our decomposition yields interpretable terms that quantify the relative contributions of position-related, token-related, and mixed effects.
By focusing on terms in this decomposition, we discover weight-based explanations of attention bias toward close tokens and attention for detokenization.

%% file: sections/1_intro.tex
\section{Introduction}\label{sec:introduction}

{
    \begin{figure*}
        \centering
        \includegraphics[width=\linewidth]{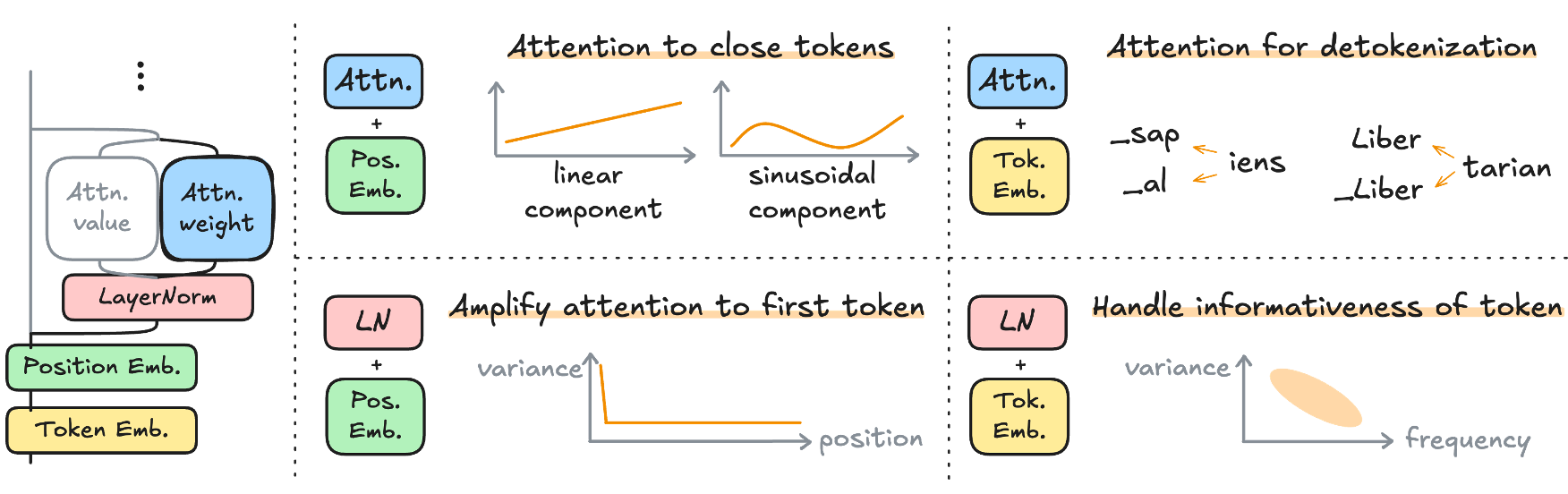}
        \caption{Focusing on the token/position embeddings, first LayerNorm layer, and the first attention layer, we conduct weight analyses and show high attention scores are assigned to close (top middle; \cref{sec:term-pos-self,sec:term-pos-compare,sec:term-positions}) \textit{and} related (top right; \cref{sec:detokenization-token}), supporting the detokenization hypothesis. We also show that the high attention score to the first token derives from LayerNorm (bottom middle; \cref{sec:wpe-var,sec:high-pos-var-at-first}). Regarding token embedding, we also discuss the relationship between token frequency and LayerNorm (bottom right; \cref{sec:wte-variance}).}
        \label{fig:fig1}
    \end{figure*}
}

Language models (LMs)~\citep{Vaswani-2017-attentionIsAllYouNeed-wv,Radford-2019-languageModelsAreUnsupervisedMultitaskLearners-wo,Brown-2020-languageModelsAreFew-shotLearners-zg,Dubey-2024-theLlama3HerdOfModels-vy,GemmaTeam-2024-gemma2ImprovingOpenLanguageModelsAtAPracticalSize-bj} operate on sequences of subword tokens 
\citep{Kudo-2018-subwordRegularizationImprovingNeuralNetworkTranslationModelsWithMultipleSubwordCandidates-tr,Sennrich-2016-neuralMachineTranslationOfRareWordsWithSubwordUnits-nq}.
Consequently, LMs encounter many words and names, e.g., ``Libertarian'', not in their natural form but split into parts such as ``Liber'' and ``tarian''.
Since such subword sequences are not necessarily linguistically meaningful, it is believed that a core function of early LM layers is to \emph{detokenize} \citep{Elhage-2022-softmaxLinearUnits-ef} sequences of subword tokens into more meaningful representations of words and names that, taken together, form the LM's \emph{inner vocabulary} \citep{Kaplan-2024-fromTokensToWordsOnTheInnerLexiconOfLlms-vi}.
This inner vocabulary contains the basic meaning representations on which subsequent stages of inference operate \citep{Lad-2024-theRemarkableRobustnessOfLlmsStagesOfInference-mx}.
However, evidence for the detokenization hypothesis has so far only been collected from empirical experiments that require selecting specific inputs and/or training probes in order to localize layers showing behavior consistent with detokenization \citep{Gurnee-2023-findingNeuronsInAHaystackCaseStudiesWithSparseProbing-vr,Kaplan-2024-fromTokensToWordsOnTheInnerLexiconOfLlms-vi}. 
Here, we take an alternative approach.
By developing a new decomposition of first-layer attention in GPT-2, we show that several important aspects of detokenization can be understood from model weights alone, without training probes or performing any inference steps (\cref{fig:fig1}).

A crucial part of detokenization is attention to tokens that, taken together, comprise a word or phrase.
Prior work has analyzed this n-gram attention aspect of detokenization \citep{Gurnee-2023-findingNeuronsInAHaystackCaseStudiesWithSparseProbing-vr,Kaplan-2024-fromTokensToWordsOnTheInnerLexiconOfLlms-vi}.
However, these analyses did not disentangle token content effects from positional biases.
Although not in the context of detokenization, \citet{Dar-2023-analyzingTransformersInEmbeddingSpace-fz} and \citet{Elhage-2021-aMathematicalFrameworkForTransformerCircuits-rr} analyze the interaction between pairs of vocabularies in the attention layer of GPT-2 and newly-trained model, respectively.
However, they do not take into account the effect of LayerNorm.
Going beyond prior work, we analyze the effect of token representation without positional information on attention weights while fully considering LayerNorm.

While such token-deriving attention is an important aspect, it is not sufficient for performing detokenization.
Another important aspect is attention to close tokens.
We conduct weight-based analysis and show that higher attention is assigned to positionally close tokens in the first layer of GPT-2~\citep{Radford-2019-languageModelsAreUnsupervisedMultitaskLearners-wo}, regardless of the input token (\cref{sec:term-pos-self,sec:term-pos-compare,sec:term-positions}).
In particular, we identify two components in the learned absolute position embeddings
The first component can be seen as a linear bias component, similar to ALiBi~\citep{Press-2022-trainShortTestLongAttentionWithLinearBiasesEnablesInputLengthExtrapolation-bb}.
The second component has a sinusoidal shape, which is reminiscent of sinusoidal \citep{Vaswani-2017-attentionIsAllYouNeed-wv} and rotary \citep{Su-2024-roformerEnhancedTransformerWithRotaryPositionEmbedding-fp} position encoding.
In superposition, these two components induce an attention bias towards close tokens.

In summary, we analyze the first attention layer in GPT-2, separating computations deriving from position embedding and token embedding.
As a whole, our results comprehensively explain and support the mechanism of detokenization in that attention layers attend to related tokens that are positioned close.

More broadly, we worked on the internal understanding of the model inference without selecting prompts and running inference.
This paper shows the first findings in this view, providing more theoretical proof of underlying mechanisms.

%% file: sections/2_background.tex
\section{Background}
Early layers of LMs are hypothesized to \textit{de-tokenize} over-segmented words and phrases.
In this section, we briefly provide the necessary background on the detokenization hypothesis.
We also discuss relevant positional encoding schemes since positional information plays an important role during detokenization.

\subsection{Detokenization}
\citet{Elhage-2022-softmaxLinearUnits-ef} introduce \textit{de-tokenization} to explain that initial layers of the language model they trained contribute to mapping multi-token or compound words to a ``semantic representation''.
For example, they found neurons responding to ``Libertarian'', which was tokenized into ``Libert'' and ``arian''.
\citet{Lad-2024-theRemarkableRobustnessOfLlmsStagesOfInference-mx} use this term as the name of the hypothesis (``detokenization hypothesis'') for the first stage of inference where language models ``integrate local context to convert raw token representations into coherent entities''.
Based on the detokenization hypothesis, \citet{Kaplan-2024-fromTokensToWordsOnTheInnerLexiconOfLlms-vi} collect multi-token words, input them to a model, and use patch scope technique \citep{Ghandeharioun-2024-patchscopesAUnifyingFrameworkForInspectingHiddenRepresentationsOfLanguageModels-iu} to inspect if original tokens can be recovered from a single hidden state between intermediate layers.
\citet{Gurnee-2023-findingNeuronsInAHaystackCaseStudiesWithSparseProbing-vr}, focusing on MLP neuron activations, train probes that distinguish n-grams, finding ``compound word neurons''.
\citet{Geva-2023-dissectingRecallOfFactualAssociationsInAuto-regressiveLanguageModels-ne}, although they do not mention ``detokenization'', report that language models build multi-token subject representations, such as ``Beats Music'', from raw token representations in the early layers.

In this work, we show some aspects of the detokenization process that can be understood just from the weights of the target LM alone, without running a single forward or backward pass.

\subsection{Positional Encoding}
Transformer-based language models explicitly use architectures that catch positional information.
\citet{Vaswani-2017-attentionIsAllYouNeed-wv} originally employed ``positional encoding'' defined by sine and cosine functions of different frequencies and added a vector representing the absolute token position at the embedding layer.
Models like BERT~\citep{Devlin-2019-bertPre-trainingOfDeepBidirectionalTransformersForLanguageUnderstanding-kc} and RoBERTa~\citep{Liu-2019-robertaARobustlyOptimizedBertPretrainingApproach-hy} use similar architecture, except the position embeddings are learned through pre-training.
\citet{Yamamoto-2023-absolutePositionEmbeddingLearnsSinusoid-likeWavesForAttentionBasedOnRelativePosition-su} report that RoBERTa, an encoder model, learns sinusoidal position embedding through training, which contributes to assigning high attention scores to close tokens.
GPT-2~\citep{Radford-2019-languageModelsAreUnsupervisedMultitaskLearners-wo} also uses learned absolute position embedding, but
they leave analyses of causal models, which showed different trends from encoder-decoder models in their analyses, to future work.

As another method for positional encoding, \mbox{ALiBi} ``biases query-key attention scores with a penalty that is proportional to their distance''~\citep{Press-2022-trainShortTestLongAttentionWithLinearBiasesEnablesInputLengthExtrapolation-bb}.
We show in \cref{sec:term-pos-self} that GPT-2 learns both the sinusoidal position bias, which relates with \citet{Yamamoto-2023-absolutePositionEmbeddingLearnsSinusoid-likeWavesForAttentionBasedOnRelativePosition-su} and the linear bias which relates with {ALiBi}.

%% file: sections/3_math_decomposition.tex
\section{Decomposing the First Attention Layer}
In this study, we select GPT-2 as our model of analysis, following prior work~\citep{Lad-2024-theRemarkableRobustnessOfLlmsStagesOfInference-mx,Dar-2023-analyzingTransformersInEmbeddingSpace-fz}, which investigates detokenization and/or attention trends.
GPT-2 is also a model that is often analyzed in prior works focusing on model interpretability in general~\citep{Geva-2023-dissectingRecallOfFactualAssociationsInAuto-regressiveLanguageModels-ne,Hanna-2023-howDoesGpt-2ComputeGreater-thanInterpretingMathematicalAbilitiesInAPre-trainedLanguageModel-hk,Conmy-2023-towardsAutomatedCircuitDiscoveryForMechanisticInterpretability-bx}.

Before conducting weight analyses, we redefine some of the weights to make analyses simple.
Specifically, focusing on LayerNorm and attention, we fold in multiple linear transformations into a single linear transformation and ignore terms that can be ignored.
We describe our analysis settings in the following subsections, showing it is mathematically equivalent to the original computation in the GPT-2 model.\footnote{We provide the detailed notation and decomposition in \cref{sec:notation,sec:decomposition_appendix}}

\subsection{Embedding}\label{sec:embedding}
In GPT-2's Embedding layer, the initial hidden state is computed based on the Token ID and the absolute position of the token.
Let the Token ID of the $i$-th token be denoted as $\text{ID}_i$, and the Token Embedding Matrix as $\bm{E} \in \mathbb{R}^{|V|\times d}$, where $|V|$ is the size of the vocabulary, and $d$ is the embedding dimension.
Additionally, GPT-2 employs absolute position embeddings, with the position embedding matrix denoted as $\bm{P} \in \mathbb{R}^{L \times d}$, where $L$ is the maximum sequence length.
The output of the Embedding layer at position $i$, $\bm{x}_i$, is thus represented as follows:%
\begin{equation}
    \bm{x}_i = \bm{e}_{\text{ID}_i} + \bm{p}_i
\end{equation}

\subsection{Layer Normalization}\label{sec:layernorm}

The Transformer architecture applies layer normalization at various points.
The specific form of layer normalization used in GPT-2, namely LayerNorm can be expressed~\footnote{We underline symbols from the original formulation to distinguish them from symbols in our reformulation in \ref{sec:redefinubg-ln-attn}} as follows:%
\begin{align}
    &\underline{\text{LN}}(\bm{x}) := \frac{\bm{x}}{\sigma(\bm{x})}\left(\bm{I}-\frac{1}{d}\bm{1}^\top\bm{1}\right)\diag(\bm\gamma)+ \bm{\beta} \label{eq:ln_decomposed}\\
    &\sigma(\bm{x}) := \sqrt{\var(\bm{x})+\epsilon}\label{eq:sigma}
\end{align}
where $\epsilon$ is a small constant added to prevent division by zero, and $\bm{\gamma}, \bm{\beta} \in \mathbb{R}^{d}$ are learnable parameters.
Thus, once division by $\sigma(\bm{x})$ has been computed, LayerNorm is simply an affine transformation.
Originally introduced to stabilize and speed up model training  \citep{Ba-2016-layerNormalization-jh}, from the perspective of interpretability, this component is mainly viewed as a nuisance factor that complicates analysis due to the appearance of the variance term $\var(\bm{x})$ taken over the hidden state $\bm{x}$.
While interpretability work has dealt with LayerNorm by ``folding'' the affine transformation part into other components~\citep{Nanda-2023-transformerlensALibraryForMechanisticInterpretabilityOfGpt-styleLanguageModels-un}, ignoring it~\citep{Dar-2023-analyzingTransformersInEmbeddingSpace-fz}, or removing it altogether \citep{Heimersheim-2024-youCanRemoveGpt2sLayernormByFine-tuning-rc}, we will show that the LayerNorm variance plays an important role by in effect acting as an if-then condition on token positions.

\subsection{Attention}\label{sec:attention}
The role of attention is to dynamically mix contextual information into the representation of the current token.
In a causal model, given current position $i$ and the context information $\bm{X}$, the attention layer with $H$ heads performs the following computation:%
\begin{equation}
    \text{ATTN}(i, \bm{X}) := \sum_{h=1}^H \sum_{j=1}^i \alpha_{i, j, h} \bm{v}_h(\bm{x}_j)\bm{W}^O_h + \bm{b}^O\label{eq:attn_original}
\end{equation}
where $\bm{v}_h(\bm{x}_j)$ is the value vector of dimension $d' = d/H$ associated with token representation $\bm{x}_j$, matrix $\bm{W}^O$ and vector $\bm{b}^O$ are the weight and bias of the affine output transformation, and the attention weights $\alpha_{i, j, h}$ from token position $i$ to $j$ in head $h$ with $\sum_j \alpha_{i, j, h} = 1$ are given by:%
\begin{equation}
\alpha_{i, j, h} := \underset{\bm{x}_j \in \bm{X}, j \leq i}{\text{softmax}}\left(s_{i, j, h}/\sqrt{d'}\right)\label{eq:attn_alpha}
\end{equation}%
Here, $s_{i, j, h}$ are unnormalized attention scores:
\begin{equation}
    \underline{s_{i, j, h}} := \bm{q}_h(\bm{x}_i)\bm{k}_h(\bm{x}_j)^\top\label{eq:attn_s_original}
\end{equation}%
where $\bm{q}_h$ and $\bm{k}_h$ are affine transformation from an input $\bm{x}$ to a query vector and key vector:%
\begin{align}
    \bm{q}_h(\bm{x}) &:= \bm{x}\bm{W}_h^Q + \bm{b}_h^Q\label{eq:attn_affine_original}\\
    \bm{k}_h(\bm{x}) &:= \bm{x}\bm{W}_h^K + \bm{b}_h^K
\end{align}%

\subsection{Reformulating LayerNorm and Attention}\label{sec:redefinubg-ln-attn}
From \cref{eq:attn_original,eq:attn_alpha}, it can be observed that in the computation of ATTN, input $\bm{x}$ are always first subjected to an affine transformation.
Because the input to ATTN is the output of LN, the linear part of \underline{LN} can be absorbed into the affine transformations for the $\bm{q}_h$, $\bm{k}_h$ and $\bm{v}_h$ functions in ATTN.
Therefore, instead of \cref{eq:ln_decomposed}, we redefine LN:%
\begin{equation}
    \text{LN}(\bm{x}) := \bm{x} / \sigma(\bm{x})\label{eq:ln-redefined}
\end{equation}%
We also redefine the affine transformation that is applied to $\bm{q}_h$ in \cref{eq:attn_affine_original}, with analogous redifintions for $\bm{k}_h$ and $\bm{v}_h$:%
\begin{align}
     \bm{W}^Q_h &:= \left(\bm{I}-\frac{1}{d}\bm{1}^\top\bm{1}\right)\diag(\bm\gamma)\underline{\bm{W}}^Q_h \\
    \bm{b}^Q_h& := \bm{\beta}\underline{\bm{W}^Q_h} + \underline{\bm{b}}^Q_h
\end{align}%

Next, we reformulate the unnormalized attention scores $\underline{s_{i, j, h}}$ defined in \cref{eq:attn_s_original}:%
\begin{align}
    \underline{s_{i, j, h}}=\bm{q}_h(\bm{x}_i)\bm{W}^{K\top}_h \bm{x}_j^\top+\bm{q}_h(\bm{x}_i)\bm{b}^{K\top}_h\label{eq:attn_with_bk}
\end{align}
Note that in \cref{eq:attn_alpha}, the softmax is applied over token positions $j$, while the second term in \cref{eq:attn_with_bk} does not depend on $j$, i.e., when computing the attention from token position $i$ to other tokens, this term is a constant.
Since the softmax is invariant to the addition of a constant, this term can be ignored in analysis, and by extension, this means that the bias $\bm{b}_h^K$ is actually completely meaningless.
By further expanding \cref{eq:attn_with_bk}, we obtain:%
\begin{equation}
    s_{i, j, h} := \bm{x}_i\bm{W}^{QK}_h\bm{x}_j^\top+\bm{b}^{QK}_h\bm{x}_j^\top \label{eq:attn_s_redefined}
\end{equation}
with
\begin{align}
    \bm{W}^{QK}_h &:= \bm{W}^{Q}_h\bm{W}^{K\top}_h\\
    \bm{b}^{QK}_h &:= \bm{b}^{Q}_h\bm{W}^{K\top}_h
\end{align}
This reformulation of the attention score computation is the basis for our decomposition, which we will perform following subsection.

\subsection{Decomposition}
The first term in \cref{eq:attn_s_redefined} depends on the hidden state of the present token $\bm{x}_i$ and the hidden state of the past token $\bm{x}_j$.
That is, this term becomes large when the two hidden states $\bm{x}_i$ and $\bm{x}_j$ are similar under the linear projection $\bm{W}^{QK}_h$, which is why we will call this term the ``comparison term''.
The second term in \cref{eq:attn_s_redefined} depends only on the hidden state of the past token $\bm{x}_j$ and is independent of the present token $i$.
A large $\bm{b}^{QK}_h\bm{x}_j^\top$ value means, figuratively, that token $j$ self-asserts its relevance regardless of context, which is why call this term the ``self-assertion term''.

So far we have ignored positional encoding.
The input to the first attention layer consists of the token embedding $\bm{e}_{\text{ID}_i}$ and position embedding $\bm{p}_i$:%
    \begin{align}
        \bm{x}_i := \frac{\bm{e}_{\text{ID}_i} + \bm{p}_i}{\sigma(\bm{e}_{\text{ID}_i} + \bm{p}_i)}\label{eq:token_pos_emb}
    \end{align}%
By plugging \cref{eq:token_pos_emb} into \cref{eq:attn_s_redefined} and expanding, we obtain a decomposition of the first layer's attention scores into six terms:
\begin{align}
    s_{i, j, h} 
    &=
    \blueboxannot{\term[i, j, h]{ee}}{\frac{\bm{e}_{\textsc{id}_i}\bm{W}_h^{QK}\bm{e}_{\textsc{id}_j}^\top}{\sigma_i\sigma_j}}
    +
    \blueboxannot{\term[i, j, h]{pp}}{\frac{\bm{p}_i\bm{W}_h^{QK}\bm{p}_j^\top}{\sigma_i\sigma_j}}\nonumber\\[17pt]
    &\phantom{= }
    +
    \blueboxannot{\term[i, j, h]{pe}}{\frac{\bm{p}_{i}\bm{W}_h^{QK}\bm{e}_{\textsc{id}_j}^\top}{\sigma_i\sigma_j}}
    +
    \blueboxannot{\term[i, j, h]{ep}}{\frac{\bm{e}_{\textsc{id}_i}\bm{W}_h^{QK}\bm{p}_{j}^\top}{\sigma_i\sigma_j}}\nonumber\\[17pt]
    &\phantom{= }
    +
    \blueboxannot{\term[j, h]{e}}
    {\frac{\bm{b}_h^{QK}\bm{e}_{\textsc{id}_j}^\top}{\sigma_j}}
    +
    \blueboxannot{\term[j, h]{p}}{\frac{\bm{b}_h^{QK}\bm{p}_j^\top}{\sigma_j}\label{eq:attn_s_6terms}}
    \nonumber\\
\end{align}%
where $\sigma_i := \sigma(\bm{e}_{\text{ID}_i} + \bm{p}_i)\label{eq:sigma-i}$.
For brevity, we will refer to each of the six terms in \cref{eq:attn_s_6terms} using the underset blue alias.

With this decomposition in place, we are now ready to analyze specific terms.
Starting with the token comparison term \term{ee}, we will show that it can be understood as representing token-token affinities and use it to identfy detokenization  heads (\cref{sec:detokenization-token}).
Then we show how the two purely position-related terms \term{p} and \term{pp} contribute to detokenization by biasing attention towards preceding tokens (\cref{sec:detokenization-position}). Finally, we analyze the remaining three terms, which are not directly related to detokenization, in \cref{sec:misc}.

%% file: sections/4_token_and_detok.tex
\section{Detokenization and Token Affinity}\label{sec:detokenization-token}
\begin{figure*}[t]
    \centering
    \includegraphics[width=\linewidth]{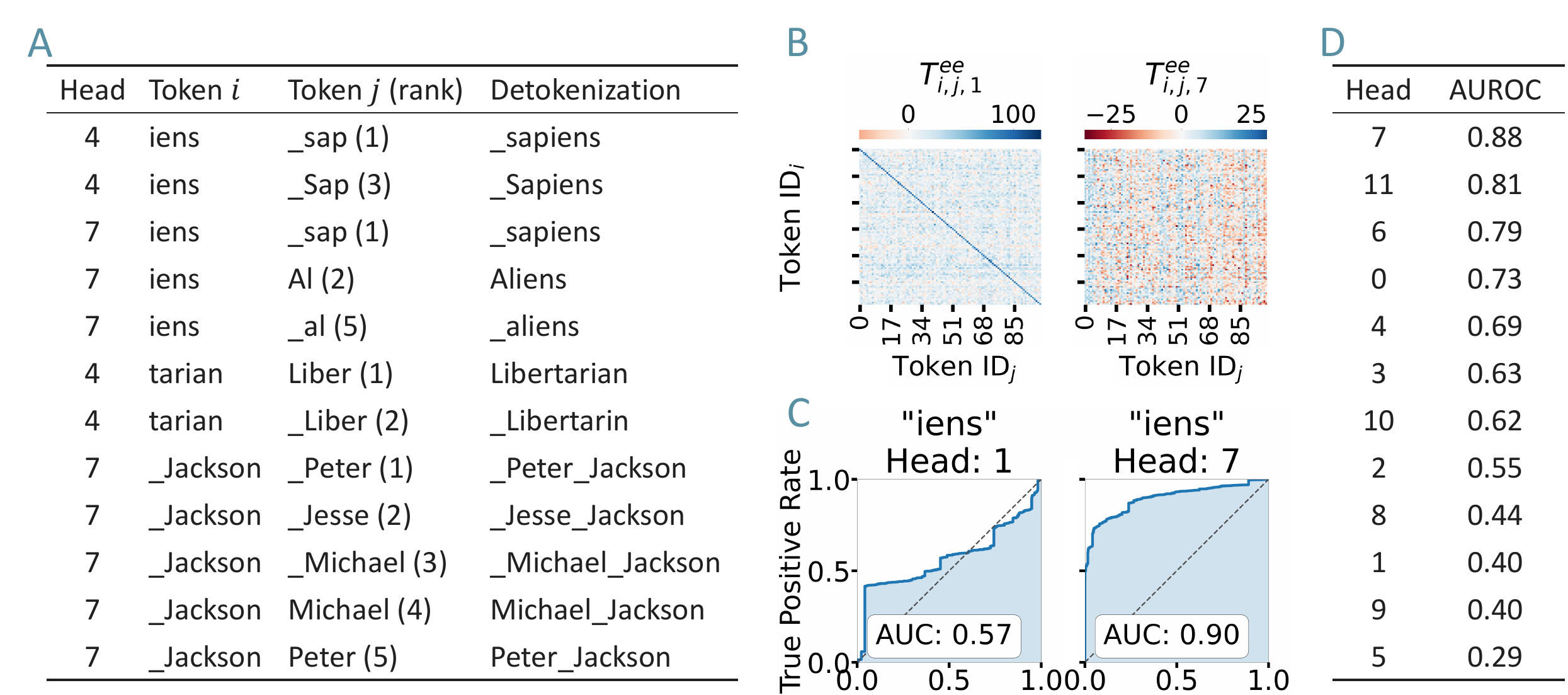}
    \caption{\textbf{\textcolor{pptblue}{A}}: Examples of support for detokenization. When the current position token is ``iens'', the past token that yields the largest \term{ee} value (=Rank 1) is ``\_sap'' in head\#4 and head\#7. \textbf{\textcolor{pptblue}{B}}: Heatmap of \term{ee} for head\#7 and head\#1. Tokens are randomly sampled from the vocabulary for visualization. \textbf{\textcolor{pptblue}{C}}: ROC of head\#7 and head\#1 when token $i$ is ``iens''. \textbf{\textcolor{pptblue}{D}}: Average AUROC for each head. Heads with high AUROC values contribute to the reconstruction of bi-grams, consequently contributing to detokenization.}
    \label{fig:detokenization-token}
\end{figure*}

The token comparison term \term{ee} is highly relevant to detokenization because its numerator $\bm{e}_{\textsc{id}_i}\bm{W}_h^{QK}\bm{e}_{\textsc{id}_j}^\top$ compares the embedding of the current token $\bm{e}_{\textsc{id}_i}$ and the past token $\bm{e}_{\textsc{id}_j}$ through the linear transformation $\bm{W}_h^{QK}$.
A large \term{ee} value means that the source token is biased to pay high attention to the target token.

In this section, we show examples of detokenization performed by \term{ee} (\cref{sec:detokenization-example}) and investigate which heads actually contribute to detokenization (\cref{sec:detokenization-head}).

\input{sections/fig_detokenization_positions}

\subsection{Examples of 
Detokenization}\label{sec:detokenization-example}
Here, we show some examples of detokenization, where attention heads assign high \term{ee} value to tokens that form words or two-token entities together with source (i.e. current) token.

First, we compiled some words or entities that are split into two tokens by the GPT-2 tokenizer (e.g., ``sapiens'').
For each instance, we fix the second token (``iens'') id as $\text{ID}_i$ and compute \term{ee} against all $\text{ID}_j$ in vocabulary $V$ in all heads.
In \cref{fig:detokenization-token}-\textbf{\textcolor{pptblue}{A}}, we show pairs of $\text{ID}_j$ and $\text{ID}_i$ with high \term{ee} that together form meaningful sequences.
For example,  when $\text{ID}_i$ is ``iens'', token ``\_sap'' and ``Al'' yields the two largest \term{ee} scores among 50,257 subwords in head\#7, detokenizing ``\_sapiens'' and ``Aliens''.
In Appendix \cref{tab:detokenization-all}, we show other examples of \term{ee} contributing to detokenizing words, people's names, or       chemical substances.

\subsection{Which Heads Perform Detokenization?}\label{sec:detokenization-head}

Before conducting a detailed analysis, we visualize the \term{ee} matrix for each of the 12 attention heads as a heatmap to gain an overview (\cref{fig:detokenization-token}-\textbf{\textcolor{pptblue}{B}}, \cref{fig:compare-tok-scores-all}).
This visualization reveals that the heads can be broadly divided into two categories: those with diagonal lines in the heatmap and those without.
Head\#1, with scores in the range (-47, 127), has a clear diagonal line, indicating that the attention score deriving from \term{ee} is the highest when the target token is identical to the current past token.
Head\#7 on the other hand, does not show such a diagonal line.
Instead, scores in range (-35, 31) are distributed across the heatmap.

What do these two trends suggest?
Detokenization typically requires attending to different tokens -- for example, paying attention from ``iens'' to ``sap'' to reconstruct the word ``sapiens,'' rather than attending from ``iens'' to ``iens.''
Therefore, heads that exhibit clear diagonal lines, such as head\#1, are less likely to contribute to detokenization compared to heads with more dispersed attention patterns, such as head\#7.

Next, we investigate the degree to which each attention head contributes to detokenization by inspecting the relationship between the scores of \term{ee} and bi-gram frequency.

For a fixed token at position $i$, we first compute \term{ee} against all 50,257 tokens in the vocabulary for GPT-2.
Using bi-gram counts of OpenWebText Corpus~\citep{Gokaslan-2019-openwebtextCorpus-bf}, we compute AUROC with True Positive Rate defined as the proportion of bi-gram counts with \term{ee} above a threshold.
\cref{fig:detokenization-token}-\textbf{\textcolor{pptblue}{C}} shows the Recall curve for head\#7 and head\#1 when the token at position $i$ is fixed to ``iens'' or ``Jackson''.
High AUC (Area Under Curve) indicates that \term{ee} with high scores are likely to reconstruct frequent bi-grams.
By computing AUC for all position $i$ tokens and taking the average, we quantify how each attention head contributes to reconstructing bi-grams.
\cref{fig:detokenization-token}-\textbf{\textcolor{pptblue}{D}}  shows the result, showing largest AUROC in head\#7 which was also observed in \cref{fig:detokenization-token}-\textbf{\textcolor{pptblue}{A}} and \cref{tab:detokenization-all}, supporting contribution to detokenization.

%% file: sections/fig_detokenization_positions.tex
\begin{figure*}[t]
    \centering
    \includegraphics[width=\linewidth]{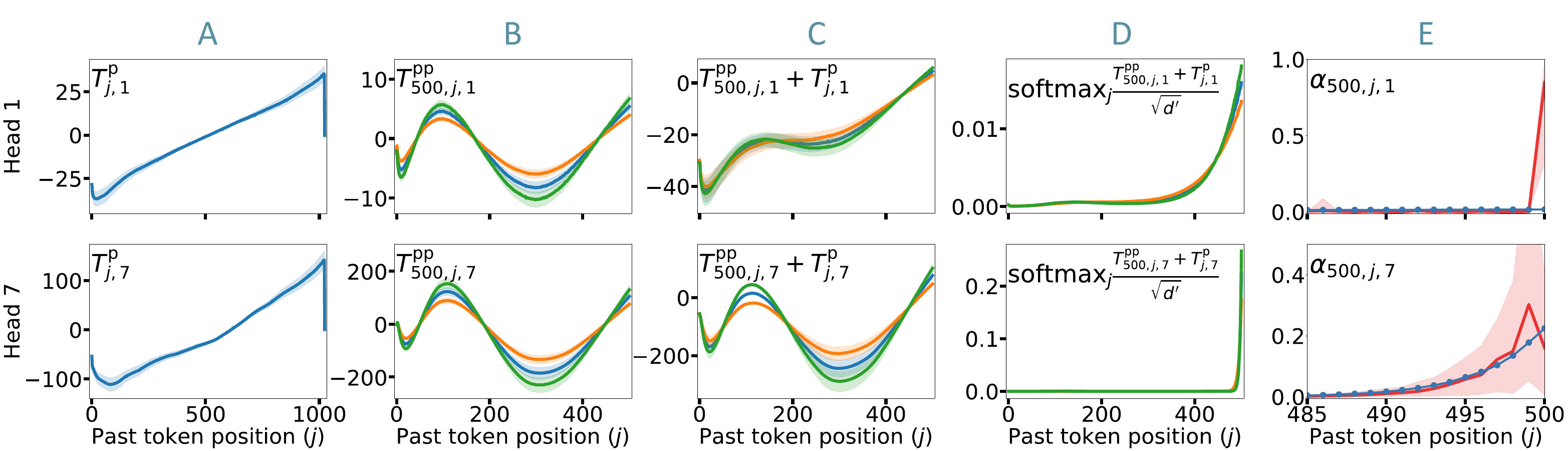}
    \caption{\textbf{{\textcolor{pptblue}{A}}}: \term[j]{p} for all context token position $j$ for head\#1 (top) and \#7 (bottom). The shaded area represents the variance of the term deriving from $e_{\text{ID}_i}$ Y-axis titles in this, and subsequent figures are inset for readability. \textbf{\textcolor{pptblue}{B}}: \term[500, j]{pp}$={\bm{p}_i\bm{W}_h^{QK}\bm{p}_j^\top}/{\sigma_i\sigma_j}$ for context token position $j\leq 500$. The \textcolor{snsblue}{blue}, \textcolor{snsgreen}{green}, and \textcolor{orange}{orange} lines show scores with mean, maximum, and minimum standard deviation for $\sigma_j$: \textcolor{snsblue}{$\sigma_j = \frac{1}{|V|}\sum_{\text{ID}}\sigma(\bm{e}_{\text{ID}} + \bm{p}_j)$}, \textcolor{snsgreen}{$\sigma_j = \text{max}_{\text{ID}} \sigma(\bm{e}_{\text{ID}} + \bm{p}_j)$}, and \textcolor{orange}{$\sigma_j = \text{min}_{\text{ID}} \sigma(\bm{e}_{\text{ID}} + \bm{p}_j)$}, respectively. \textbf{\textcolor{pptblue}{C,D}}: Sum of \term[500, j]{pp} and \term[j]{p} and its result after taking softmax with a temperature of $\sqrt{d'}$. \textbf{\textcolor{pptblue}{E}}: Empirical attention weights aggregated over texts in OpenWebText Corpus when present token position $i=500$ focusing on the last few $j$ positions. The \textcolor{snsred}{red line and area} show the empirically observed weights and the \textcolor{snsblue}{blue line} corresponds with the blue lines in \textbf{\textcolor{pptblue}{D}}.}
    \label{fig:detokenization-position}
\end{figure*}

%% file: sections/5_position_and_detok.tex
\section{Detokenization and Token Positions}\label{sec:detokenization-position}

Another important aspect is attention to close tokens.
We conduct weight-based analysis on \term{p} (\cref{sec:term-pos-self}) and \term{pp} (\cref{sec:term-pos-compare}) and identify two different trends: one linear and one sinusoidal.
Then we show in \cref{sec:term-positions} show that the sum of the two components biases attention to positionally close tokens.
Finally, we verify the positional bias empirically in \cref{sec:empirical-tp-tpp-verification}.

\subsection{Linear Self-assertion Term (\term{p})}\label{sec:term-pos-self}

We visualize the position-deriving self-assertion term \term{p} for head\#1 and head\#7 in \cref{fig:detokenization-position}-\textbf{\textcolor{pptblue}{A}} \footnote{Results for other heads are in appendix \cref{fig:self-pos-score-all}.}.
Since GPT-2 is a causal model, the causal mask ensures that no context after the present token is referenced.
For example, when $i=200$, the part of \cref{fig:detokenization-position}-\textbf{\textcolor{pptblue}{A}} to the right of $j=200$ is ignored.
In the range $0 \leq j \leq 200$, since the self-assertion term is monotonically increasing with respect to $j$, we can observe that high attention scores are assigned to tokens that are relatively positionally close.
The same applies for almost all $i$,
thereby constituting a bias towards high attention scores to nearby tokens.~\footnote{We show analyses of the exceptional behavior shown in the first and last few positions in \cref{sec:wpe-var}.}

\subsection{Sinusoidal Comparison Term (\term{pp})}\label{sec:term-pos-compare}

We visualize the position-deriving comparison term \term{pp} for head\#1 and head\#7 in \cref{fig:detokenization-position}-\textbf{\textcolor{pptblue}{B}}  when $i=500$\footnote{Results for other heads and $i$ are in appendix \cref{fig:compare-pos-score-all}.}.
A notable difference from \cref{fig:detokenization-position}-\textbf{\textcolor{pptblue}{A}} is that this term exhibits undulating patterns, which could be related to the observations by \citet{Yamamoto-2023-absolutePositionEmbeddingLearnsSinusoid-likeWavesForAttentionBasedOnRelativePosition-su}.
However, what is similar is that this term also assigns high attention scores to close tokens, contributing to high attention score on nearby tokens.

\subsection{Sum of \term{pp} and \term{p}}\label{sec:term-positions}

\cref{fig:detokenization-position}-\textbf{\textcolor{pptblue}{C}} shows the sum of the two terms deriving from position embedding\footnote{Results for other heads and $i$ are in Appendix \cref{fig:both-pos-scores-all}}.
The combination of the monotonic increase in the \term{p} and the sinusoidal component from the \term{pp} leads to high attention scores being assigned to tokens positioned nearby.

The result after applying softmax function to the sum of the two terms, is shown in \cref{fig:detokenization-position}-\textbf{\textcolor{pptblue}{D}}.
It indicates that the attention weight is concentrated on nearby tokens, and the sinusoidal component originating from the \term{pp} is almost entirely suppressed.

\subsection{Empirical Verification}\label{sec:empirical-tp-tpp-verification}

To verify whether the observations through the weight analyses hold true empirically, we use natural language texts from OpenWebText Corpus~\citep{Gokaslan-2019-openwebtextCorpus-bf} and obtain $\alpha_{i, j, h}$ from the first attention layer.
\cref{fig:detokenization-position}-\textbf{\textcolor{pptblue}{E}} shows the results when $i=500$.
Head\#7 shows a pattern similar to the plot in \cref{fig:detokenization-position}-\textbf{\textcolor{pptblue}{D}} (also shown in blue line), with high attention to the nearest tokens.
In contrast, head\#1 predominantly attends to itself (Blue line, $j=500$), which differs from the red line (also shown in \cref{fig:both-pos-scores-all}).
The discrepancy can be attributed to the dominance of \term{ee}.
We showed in \cref{fig:detokenization-token}-\textbf{\textcolor{pptblue}{B}} that head\#1 assigns high attention scores to the identical token, with scores in range (-50, 100).
Furthermore, from \cref{fig:detokenization-position}-\textbf{\textcolor{pptblue}{B}}, it can be observed that head\#1 assigns scores within a narrower range (-40, 0) across broad positions.

%% file: sections/6_followup.tex
\section{Followup Experiments}\label{sec:misc}
\subsection{Contribution of the Six Terms to Attention Weights}\label{sec:empirical-6terms}
{
    \begin{figure}
        \centering
        \includegraphics[width=\linewidthcoef\linewidth]{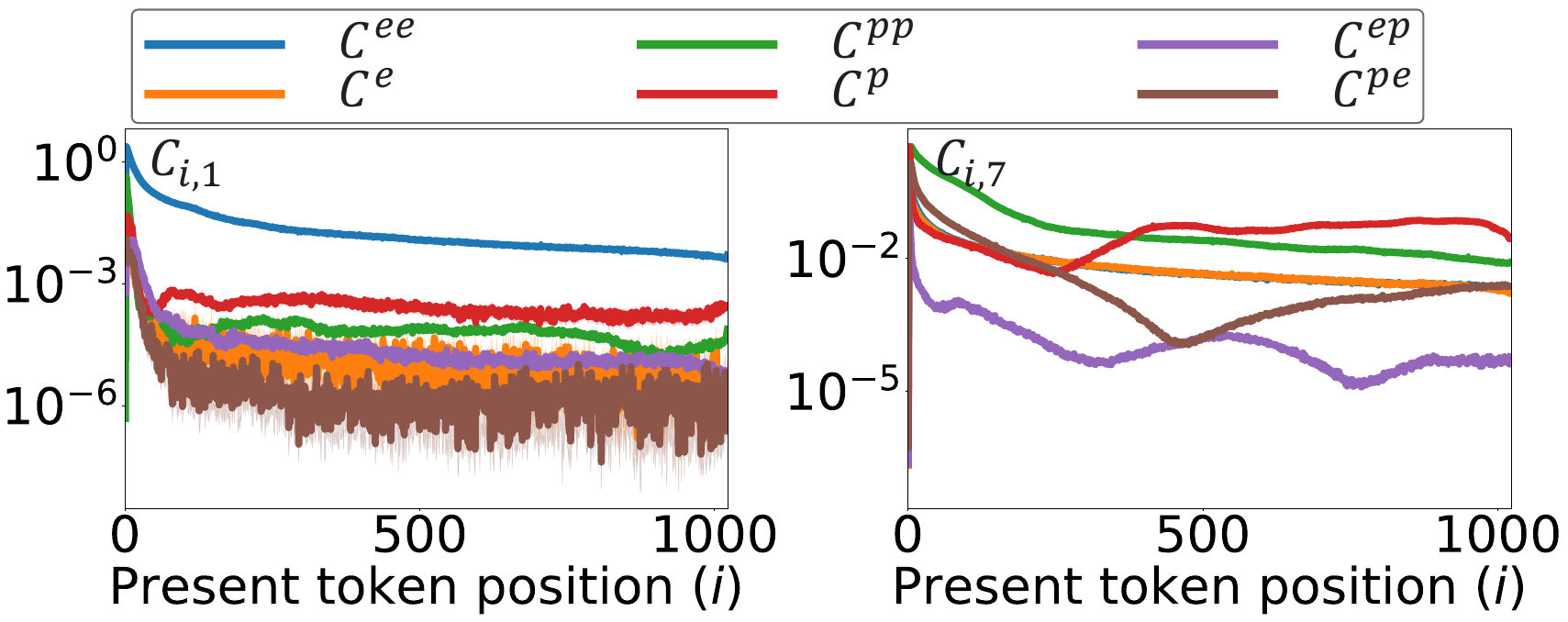}
        \caption{Contribution of the 6 terms in \cref{eq:attn_s_6terms} for each current token position $i$ for head\#1 and \#7}
        \label{fig:term-contribution}
    \end{figure}
}
In \cref{sec:detokenization-token,sec:detokenization-position}, we conducted weight analyses of \term{ee}, \term{p}, and \term{pp}.
Here, we inspect the contribution of all 6 terms in \cref{eq:attn_s_6terms} to check if we are missing any crucial terms.
We use KL-Divergence as a metric.
For example, we define the contribution of the \term{ee}, $c^{\text{ee}}$ as follows:
\begin{align}
    c^{\text{ee}}_{i, h}&=D_{\text{KL}}(P_{i, h}||Q_{i, h})\\
    Q_{i, h} &=
    \begin{bmatrix}
        \alpha_{i, 0, h} & \cdots & \alpha_{i, i, h}
    \end{bmatrix}\\
    P_{i, h} &=
    \begin{bmatrix}
        \alpha'_{i, 0, h} & \cdots & \alpha'_{i, i, h}
    \end{bmatrix}\\
    \alpha'_{i, j, h} &= \underset{\bm{x}_j \in \bm{X}, j \leq i}{\text{softmax}}\left(\frac{s_{i, j, h} - T^{\text{ee}}_{i, j, h} }{\sqrt{d'}}\right)
\end{align}

\cref{fig:term-contribution} shows that the three terms analyzed in \cref{sec:detokenization-position,sec:detokenization-token} have relatively high contributions to the attention weight computation.
From the three remaining terms, \term{pe} and \term{ep}, contribute less.
However, \term{e} shows a high contribution.
We analyze this term in the following section.

\subsection{Token Self-assertion Term (\term{e})}\label{sec:term-tok-self}
{
    \begin{figure}
        \centering
        \includegraphics[width=\linewidthcoef\linewidth]{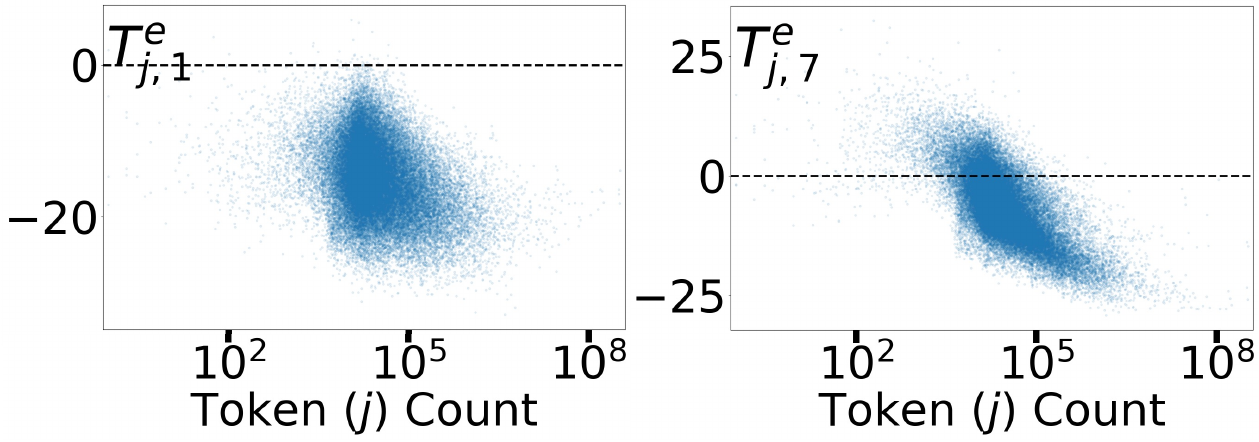}
        \caption{Relation between token frequency and \term{e} for head\#1 and head\#7.}
        \label{fig:self-tok-score}
    \end{figure}
}
\term{e} could be interpreted as a bias term for which token to attend to.
Here, we use token frequency as a proxy for the informativeness of a token and plot the relationship between token count and \term{e} in \cref{fig:self-tok-score}\footnote{We show the results for other heads in \cref{fig:self-tok-scores-all}.}.
We used the OpenWebText Corpus~\citep{Gokaslan-2019-openwebtextCorpus-bf} to get token counts.
The results show that several heads exhibit a high correlation with token frequency.
For example, the Spearman correlation coefficient between \term{e} and token frequency for head\#7 is $-0.68$, while for head\#1, it is $-0.31$.

\subsection{Variance of token embeddings}\label{sec:wte-variance}

{
    \begin{figure}[t] 
        \centering
        \includegraphics[width=\linewidthcoef\linewidth]{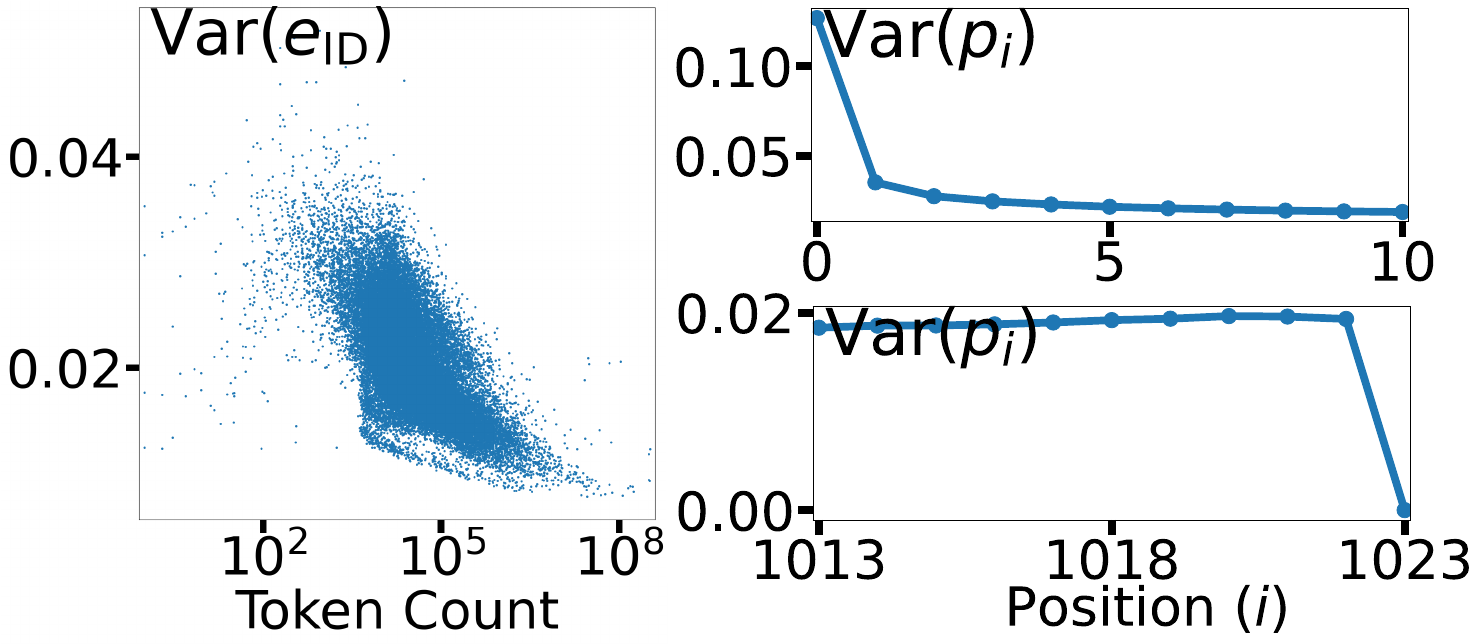}
        \caption{\textbf{Left}: Relationship between the variance of each token embedding and their corpus counts. \textbf{Right}: Variance of all first 10 (top), and last 10 (bottom) position embeddings.}
        \label{fig:variance}
    \end{figure}
}

The variance of token embeddings plays a critical role in LayerNorm (\cref{eq:ln-redefined}).
\cref{fig:variance}-left shows the relationship between token count and variance of the corresponding embedding vector.
The Spearman correlation coefficient of this relation is $-0.63$, indicating the variance of rare tokens is high.
The variance of the norm of each token embedding vector is $0.19$.
After dividing by the root of the variance, the variance becomes $0.00$.
That is, when applying LayerNorm directly to the token embeddings it has the effect of equalizing the norms of all token embeddings\footnote{Constant variance of position embedding(\cref{sec:wpe-var}) and small mean absolute covariance of position embedding and token embedding (46 times smaller than the mean variance of token embedding) supports the validity of this observation.}.
\citet{Oyama-2023-normOfWordEmbeddingEncodesInformationGain-lb} shows that the norm of word embedding encodes information gain.
However, as shown in \cref{fig:self-tok-score}, even after passing through LayerNorm, information on token count can be extracted.
Though it may depend on the definition of information gain, we speculate that something akin to information gain is encoded in places other than the norm.

\subsection{Variance of Position Embeddings}\label{sec:wpe-var}

We visualize the variance of the position embeddings, which plays a critical role in LayerNorm when visualizing over context position $j$ as suggested by \cref{eq:ln-redefined,eq:sigma,eq:sigma-i}, in \cref{fig:variance}-right.
For most positions, the variance remains constant.
However, there are two exceptions: the variance for the first token is exceptionally high, and the variance for the last token is significantly low. 
As a result, the variance of the position embeddings takes the shape of a step function.

The large variance at position 0, combined with the negative attention score without LayerNorm at $j=0$, leads to an exceptionally high attention score.
It creates a non-linear function implementing an ``if-then'' condition: if the attention is directed towards the first token of the context, the attention weight is amplified.
We discuss why such ``switch'' is implemented and why the variance of the last position takes exceptional value in \cref{sec:high-pos-var-at-first}.

\paragraph{Exceptionally High Variance at First Position}\label{sec:high-pos-var-at-first}
\citet{Gurnee-2024-universalNeuronsInGpt2LanguageModels-kh} reports that the norm of the value vector for the BOS token of GPT-2-medium is 19.4 times smaller than the average for other tokens and looks for neurons that may utilize the BOS token as \textit{attention sink}~\citep{Xiao-2023-efficientStreamingLanguageModelsWithAttentionSinks-xn} via ablation studies.
1We propose an alternative interpretation of these findings: by exceptionally increasing the variance of the first position embedding, the first token is used as an attention sink.
However, the attention sink may not be used depending on the context length and the head.
In fact, while \cref{fig:detokenization-position}-\textbf{\textcolor{pptdarkblue}{A}} has a high attention score assigned to the first token, \cref{fig:detokenization-position}-\textbf{\textcolor{pptdarkblue}{D}} has the score suppressed\footnote{Refer to \cref{fig:both-pos-scores-all} for other variations.}.
It can be interpreted that the mechanism is embedded in the LayerNorm term common to all heads and all current positions $i$ because, as the context length increases, the scores for nearby tokens, mainly due to \term{p}, exceed the scores for the first token, rendering the attention sink possibly unnecessary and ignorable by the softmax function.

\paragraph{Exceptionally Low Variance at Last Position}\label{sec:low-pos-var-at-last}
{
    \begin{figure}
        \centering
        \includegraphics[width=\linewidthcoef\linewidth]{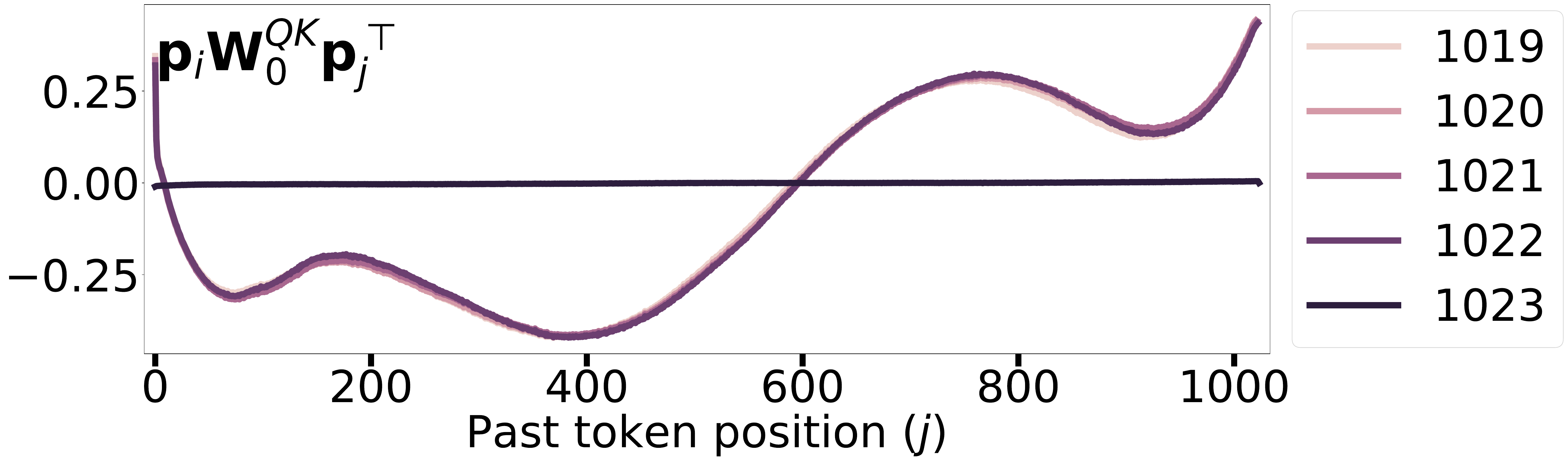}
        \caption{\term[i,j,0]{pp} without LayerNorm for $1019 \leq i \leq 1023$. When the current position $i$ is 1023, the maximum input length of GPT-2, the attention scores show a distinct outlier behavior.}
        \label{fig:wpe-last-undertrained}
    \end{figure}
}
A plausible explanation for the irregularity observed at the last position in \cref{fig:variance}-bottom-right is undertraining.
\cref{fig:wpe-last-undertrained} visualizes the \term{pp} without LayerNorm for the final positions, specifically for $1019\leq i \leq 1023$.
While the plots largely overlap for $1019 \leq i \leq 1022$, a distinct deviation can be seen at $i=1023$, where the score remains consistently flat around zero (black line).
We suspect that this phenomenon occurs because the maximum length for prompts accepted by GPT-2 is 1024 tokens, and the model was not trained with the loss for the 1025th token.

%% file: sections/7_conclusions.tex
\section{Conclusions}
We analyzed the first attention layer of GPT-2 to investigate the detokenization mechanism.
Considering detokenization to be a phenomenon that occurs when deeply related tokens are close together, we conducted analyses that separate token embedding and position embedding to provide multiple theoretical supports for detokenization.
First, we showed that the self-assertion attention term and comparison term derived from position embedding contribute to assigning high attention to relatively close tokens.
Furthermore, we suggested that position embedding and LayerNorm are deeply related to the phenomenon of high attention being assigned to the first token.
Regarding the relevance between tokens, we showed that the comparison term deriving from token embedding contributes to this, and obtained results that are generally consistent with existing research.

%% file: sections/8_limitations.tex
\section*{Limitations}
In this study, we analyzed the weights learned by the first layer of GPT-2.

The detokenization mechanism itself does not occur only in the first layer, thus further analysis of other early layers may be necessary for a deeper analysis.
However, by analyzing a small version of GPT-2, a 12-layer model, we may have captured the main mechanism of detokenization even with only one layer.

The analysis we conducted, which separates the attention weight calculation into token embedding and position embedding, was possible and meaningful because GPT-2 adopts Learned Absolute Position Embedding.
For other models that use ALiBi or Rotary Position Embedding, it may be apparent that models attend to close tokens as the parameters in these methods are defined prior to training.
We believe our contribution lies in that we show that even without an explicit design of position bias, models can learn to attend to close tokens and contribute to detokenization from the first layer.

Furthermore, the self-assertion term is a term that appears because the transformation for calculating the key and query vectors is an affine transformation with a bias.
However, recent models such as Llama~3~\citep{Dubey-2024-theLlama3HerdOfModels-vy} and Gemma~2~\citep{GemmaTeam-2024-gemma2ImprovingOpenLanguageModelsAtAPracticalSize-bj} do not use bias terms in linear layers.
While this means analyses breaking down attention into self-assertion terms and comparison terms cannot be done on such models, we believe that our results also raise the question of whether bias terms can really be eliminated.

Regarding the choice of subject model for analysis, our ultimate goal is to gain insights into robust and efficient models.
Therefore, we believe that while the subject model must achieve a certain level of performance, it is not necessary to analyze only the SOTA models.
In addition, we believe that designing and analyzing models that are easier to interpret is one direction to take in the context of the recent discussion on the reliability of language models.

Another limitation is that our analysis explains only a part of the detokenization process. Detokenization is not just attending to close preceding tokens, but \emph{selectively} doing so, since single-token words do not require detokenization.
So the ``attend to close preceding tokens'' mechanism should be ``switched off'' for these tokens, and it is one of the future works to offer a weight-based explanation of how and where this ``switch'' is implemented.

%% file: sections/9_ethics.tex
\section*{Ethics Statement}
Our analyses involved a pretrained language model, GPT-2~\citep{Radford-2019-languageModelsAreUnsupervisedMultitaskLearners-wo} (124M parameter model from huggingface\footnote{\href{https://huggingface.co/openai-community/gpt2}{https://huggingface.co/openai-community/gpt2}}, MIT License), and corpus, OpenWebText, which may be characterized by various forms of social biases.
We analyze how GPT-2 processes natural language inputs, which is within the scope of its intended usage.

%% file: sections/9_acknowledgements.tex
\section*{Acknowledgements}
This work was supported by JST/CREST (\mbox{JPMJCR20D2}) and JST/BOOST (\mbox{JPMJBS2421}).
We would also like to thank the members of \mbox{TohokuNLP} for their frequent participation in discussions during the course of this research.

%% file: sections/9_appendix.tex
\section{Notation}\label{sec:notation}
\begin{alignat}{4}
    &\bm{E} &:= &
    \begin{bmatrix}
        \bm{e}_1\\
        \vdots\\
        \bm{e}_{|V|}
    \end{bmatrix}
    &\hspace{1em}\in &\mathbb{R}^{|V| \times d}\\
    &\bm{P} &:= &
    \begin{bmatrix}
        \bm{p}_1\\
        \vdots\\
        \bm{p}_{L}
    \end{bmatrix}
    &\hspace{1em}\in &\mathbb{R}^{L \times d}\\
    &\bm{X} &:= &
    \begin{bmatrix}
        \bm{x}_1\\
        \vdots\\
        \bm{x}_n
    \end{bmatrix}
    &\hspace{1em}\in &\mathbb{R}^{n \times d}\\
    &\bm{W}^O &:= &
    \begin{bmatrix}
        \bm{W}^O_1\\
        \vdots\\
        \bm{W}^O_H
    \end{bmatrix}
    &\hspace{1em}\in &\mathbb{R}^{d \times d}\\
    &\bm{W}^Q &:= &
    \begin{bmatrix}
        \bm{W}^Q_1 & \cdots & \bm{W}^Q_H
    \end{bmatrix}
    &\hspace{1em}\in &\mathbb{R}^{d \times d}& \label{eq:wq_split}\\
    &\bm{W}^K &:= &
    \begin{bmatrix}
        \bm{W}^K_1 & \cdots & \bm{W}^K_H
    \end{bmatrix}
    &\hspace{1em}\in &\mathbb{R}^{d \times d}& \label{eq:wk_split}\\
    &\bm{W}^V &:= &
    \begin{bmatrix}
        \bm{W}^V_1 & \cdots & \bm{W}^V_H
    \end{bmatrix}
    &\hspace{1em}\in &\mathbb{R}^{d \times d}&\label{eq:wv_split}\\
    &\bm{b}^Q &:= &
    \begin{bmatrix}
        \bm{b}^Q_1 & \cdots & \bm{b}^Q_H
    \end{bmatrix}
    &\hspace{1em}\in &\mathbb{R}^{d}& \label{eq:bq_split}\\
    &\bm{b}^K &:= &
    \begin{bmatrix}
        \bm{b}^K_1 & \cdots & \bm{b}^K_H
    \end{bmatrix}
    &\hspace{1em}\in& \mathbb{R}^{d}& \label{eq:bk_split}\\
    &\bm{b}^V &:= &
    \begin{bmatrix}
        \bm{b}^V_1 & \cdots & \bm{b}^V_H
    \end{bmatrix}
    &\hspace{1em}\in &\mathbb{R}^{d}& \\
    &\bm{I} &:= &
    \begin{bmatrix}
        1 & 0 & \cdots & 0 \\
        0 & 1 & \cdots & 0 \\
        \vdots & \vdots & \ddots & \vdots \\
        0 & 0 & \cdots & 1 \\
    \end{bmatrix}
    &\hspace{1em}\in &\mathbb{R}^{d\times d}& \\
    &\bm{1} &:= &
    \begin{bmatrix}
        1 & \cdots & 1
    \end{bmatrix}
    &\hspace{1em}\in &\mathbb{R}^{d}
\end{alignat}

\section{Decomposing the First Attention Layer}\label{sec:decomposition_appendix}

\subsection{Layer Normalizaion}
Layer Normalization can be expressed as follows:

\begin{alignat}{3}
    &\underline{\text{LN}}(\bm{x}) &:=&\ \frac{\bm{x}-\bm{\mu}(\bm{x})}{\sigma(\bm{x})}\odot\bm{\gamma} + \bm{\beta}&\hspace{1em}\in&\mathbb{R}^d\\
    &\bm{x} &:=&\ 
    \begin{bmatrix}
        x^{(1)} & \cdots & x^{(d)}
    \end{bmatrix}
    &\hspace{1em}\in&\mathbb{R}^d\\
    &\bm{\mu}(\bm{x}) &:=&\ m(\bm{x})\bm{1}&\hspace{1em}\in&\mathbb{R}^d\\
    &m(\bm{x}) &:=&\ \frac{1}{d}\sum_kx^{(k)}&\hspace{1em}\in&\mathbb{R}\\
    &\sigma(\bm{x}) &:=&\ \sqrt{\frac{1}{d}\sum_k^d\left(x^{(k)}-m(\bm{x})\right)^2+\epsilon}&\hspace{1em}\in&\mathbb{R}
\end{alignat}
Now, $\bm{\mu}(\bm{x})$ can be reformulated as follows:
\begin{align}
    \bm{\mu}(\bm{x})
    &=m(\bm{x})\bm{1}\\
    &=\left(\frac{1}{d}\sum_kx^{(k)}\right)\bm{1}\\
    &=\left(\frac{1}{d}\bm{x}\bm{1}^\top\right)\bm{1}\\
    &=\bm{x}\left(\frac{1}{d}\bm{1}^\top\bm{1}\right)
\end{align}
Thus \underline{LN} can be reformulated as follows.
\begin{align}
    \underline{\text{LN}}(\bm{x}) 
    &= \frac{\bm{x}-\bm{\mu}(\bm{x})}{\sigma(\bm{x})}\odot\bm{\gamma} + \bm{\beta}\\
    &= \frac{1}{\sigma(\bm{x})} \left(\bm{x}-\bm{x}\left(\frac{1}{d}\bm{1}^\top\bm{1}\right)\right) \diag{\bm\gamma}+ \bm{\beta}\\
    &= \frac{\bm{x}}{\sigma(\bm{x})}\left(\bm{I}-\frac{1}{d}\bm{1}^\top\bm{1}\right)\diag{\bm\gamma}+ \bm{\beta}
\end{align}

\subsection{Attention}
Let query, key, value transformations of each head $h$ be expressed as follows:
\begin{align}
    \bm{q}_h(\bm{x}) &:= \bm{x}\underline{\bm{W}_h^Q} + \underline{\bm{b}_h^Q}\\
    \bm{k}_h(\bm{x}) &:= \bm{x}\underline{\bm{W}_h^K} + \underline{\bm{b}_h^K}\\
    \bm{v}_h(\bm{x}) &:= \bm{x}\underline{\bm{W}_h^V} + \underline{\bm{b}_h^V}
\end{align}%
The output of Attention layer of an causal model at position $i$ can be expressed as follows:
\begin{align}
    \text{ATTN}(i, \bm{X})
        &:=\left[\text{head}_1(i, \bm{X})\hspace{0.5em}\cdots\hspace{0.5em}\text{head}_H(i, \bm{X})\right]
            \bm{W}^O + \bm{b}^O\\
        &=\sum_{h=1}^H \text{head}_h(i, \bm{X})\bm{W}^O_h + \bm{b}^O\\
        &=\sum_{h=1}^H \left(\sum_{j=1}^i \alpha_{i, j, h} \bm{v}_h(\bm{x}_j)\right)\bm{W}^O_h + \bm{b}^O\\
        &=\sum_{h=1}^H \left(\sum_{j=1}^i \alpha_{i, j, h} \bm{x}_j\bm{W}^V_h + \bm{b}^V_h\right)\bm{W}^O_h + \bm{b}^O\\
        &= \sum_{h=1}^H \sum_{j=1}^i \alpha_{i, j, h} \bm{x}_j\bm{W}^V_h\bm{W}^O_h + \bm{b}^V\bm{W}^O + \bm{b}^O\\
        &= \sum_{h=1}^H \sum_{j=1}^i \alpha_{i, j, h} \bm{x}_j\bm{W}^{VO}_h + \bm{b}^{VO}
\end{align}
where, $\alpha$ represents the attention weights $\alpha_{i, j, h}$ from token position $i$ to $j$ in head $h$, which satisfy $\sum_j \alpha_{i, j, h} = 1$ and is defined by:
\begin{align}
\alpha_{i, j, h} 
&:= \underset{\bm{x}_j \in \bm{X}, j \leq i}{\text{softmax}}\frac{\blueboxannottop{$\underline{s_{i, j, h}}$}{\bm{q}_h(\bm{x}_i)\bm{k}_h(\bm{x}_j)^\top}}{\sqrt{d'}}
\end{align}
with $s_{i, j, h}$ representing unnormalized attention scores.

\subsection{Reformulating LayerNorm and Attention}
Because the softmax function is invariant to the addition of a constant, the computation of attention score $s_{i, j, h}$ can be simplified:
\begin{align}
\nonumber\\
\alpha_{i, j, h} 
&:= \underset{\bm{x}_j \in \bm{X}, j \leq i}{\text{softmax}}\frac{\blueboxannottop{$\underline{s_{i, j, h}}$}{\bm{q}_h(\bm{x}_i)\bm{k}_h(\bm{x}_j)^\top}}{\sqrt{d'}}\\
&= \underset{\bm{x}_j \in \bm{X}, j \leq i}{\text{softmax}}\frac{\bm{q}_h(\bm{x}_i)\bm{W}^{K\top}_h \bm{x}_j^\top+\bm{q}_h(\bm{x}_i)\bm{b}^{K\top}_h}{\sqrt{d'}}\\
&= \underset{\bm{x}_j \in \bm{X}, j \leq i}{\text{softmax}}\frac{\bm{q}_h(\bm{x}_i)\bm{W}^{K\top}_h \bm{x}_j^\top}{\sqrt{d'}}\\
&= \underset{\bm{x}_j \in \bm{X}, j \leq i}{\text{softmax}}\frac{\bm{x}_i\bm{W}^{Q}_h\bm{W}^{K\top}_h\bm{x}_j^\top+\bm{b}^{Q}_h\bm{W}^{K\top}_h\bm{x}_j^\top}{\sqrt{d'}}\\[17pt]
&= \underset{\bm{x}_j \in \bm{X}, j \leq i}{\text{softmax}}\frac{\blueboxannottop{$s_{i, j, h}$}{\bm{x}_i\bm{W}^{QK}_h\bm{x}_j^\top+\bm{b}^{QK}_h\bm{x}_j^\top}}{\sqrt{d'}}
\end{align}

Meanwhile, in GPT-2, the output of a LayerNorm is fed to the corresponding Attention layer, and inputs ($\bm{x}$) are always first subjected to an affine transformation, either $\bm{q}_h(\bm{x})$,$\bm{k}_h(\bm{x})$, or $\bm{v}_h(\bm{x})$.
Combining the LayerNorm and affine transformations, we redefine LN, and the weights and biases of affine transformations $\bm{q}_h$, with analogous redifintions for $\bm{k}_h$ and $\bm{v}_h$:
\begin{align}
    \bm{q}_h\left(\underline{\text{LN}}(\bm{x})\right)
    &= \underline{\text{LN}}(\bm{x})\underline{\bm{W}_h^Q} + \underline{\bm{b}_h^Q}\\
    &= \left(\frac{\bm{x}}{\sigma(\bm{x})}\left(\bm{I}-\frac{1}{d}\bm{1}^\top\bm{1}\right)\diag{\bm\gamma}+ \bm{\beta}\right)\underline{\bm{W}_h^Q} + \underline{\bm{b}_h^Q}\\
    &= \redbox{\frac{\bm{x}}{\sigma(\bm{x})}}\greenbox{\left(\bm{I}-\frac{1}{d}\bm{1}^\top\bm{1}\right)\left(\diag{\bm\gamma}\right)\underline{\bm{W}_h^Q}} + \yellowbox{\bm{\beta}\underline{\bm{W}_h^Q} + \underline{\bm{b}_h^Q}}\\
    &= \redbox{\text{LN}(\bm{x})}\greenbox{\bm{W}_h^Q}+ \yellowbox{\bm{b}_h^Q}
\end{align}

\subsection{Decomposition of Attention Scores for the First Layer.}
At position $i$, the input to the first LayerNorm before the first Attention layer ($\bm{x}_i$) is the sum of the token embedding $\bm{e}_{\text{ID}_i}$ and position embedding $\bm{p}_i$.
Therefore, $s_{i, j, h}$ for the first Attention layer can be decomposed as follows.
\begin{align}
    s_{i, j, h}
    &= \text{LN}(\bm{x}_i)\bm{W}^{QK}_h\left(\text{LN}(\bm{x}_j)\right)^\top+\bm{b}^{QK}_h\left(\text{LN}(\bm{x}_j)\right)^\top\\
    &= \frac{\bm{e}_{\text{ID}_i}+\bm{p}_i}{\sigma(\bm{x}_i)} \bm{W}^{QK}_h \left(\frac{\bm{e}_{\text{ID}_j}+\bm{p}_j}{\sigma(\bm{x}_j)}\right)^\top + \bm{b}^{QK}_h\left(\frac{\bm{e}_{\text{ID}_j}+\bm{p}_j}{\sigma(\bm{x}_j)}\right)^\top\\
    &=
    \blueboxannot{\term[i, j, h]{ee}}{\frac{\bm{e}_{\textsc{id}_i}\bm{W}_h^{QK}\bm{e}_{\textsc{id}_j}^\top}{\sigma(\bm{x}_i)\sigma(\bm{x}_j)}}
    +
    \blueboxannot{\term[i, j, h]{ep}}{\frac{\bm{e}_{\textsc{id}_i}\bm{W}_h^{QK}\bm{p}_{j}^\top}{\sigma(\bm{x}_i)\sigma(\bm{x}_j)}}
    +
    \blueboxannot{\term[i, j, h]{pe}}{\frac{\bm{p}_{i}\bm{W}_h^{QK}\bm{e}_{\textsc{id}_j}^\top}{\sigma(\bm{x}_i)\sigma(\bm{x}_j)}}
    +
    \blueboxannot{\term[i, j, h]{pp}}{\frac{\bm{p}_i\bm{W}_h^{QK}\bm{p}_j^\top}{\sigma(\bm{x}_i)\sigma(\bm{x}_j)}}
    \nonumber\\[17pt]
    &\phantom{= }
    +
    \blueboxannot{\term[j, h]{e}}
    {\frac{\bm{b}_h^{QK}\bm{e}_{\textsc{id}_j}^\top}{\sigma(\bm{x}_j)}}
    +
    \blueboxannot{\term[j, h]{p}}{\frac{\bm{b}_h^{QK}\bm{p}_j^\top}{\sigma(\bm{x}_j)}}
    \nonumber\\[3pt]
\end{align}

\section{Qualitative Analysis of Detokenization}
We show other examples of detokenization observed from \term{ee} in \cref{tab:detokenization-all}.

\section{Visualizations for all heads}\label{sec:figs-allheads}
\begin{figure}[p]
    \centering
    \includegraphics[width=\linewidth]{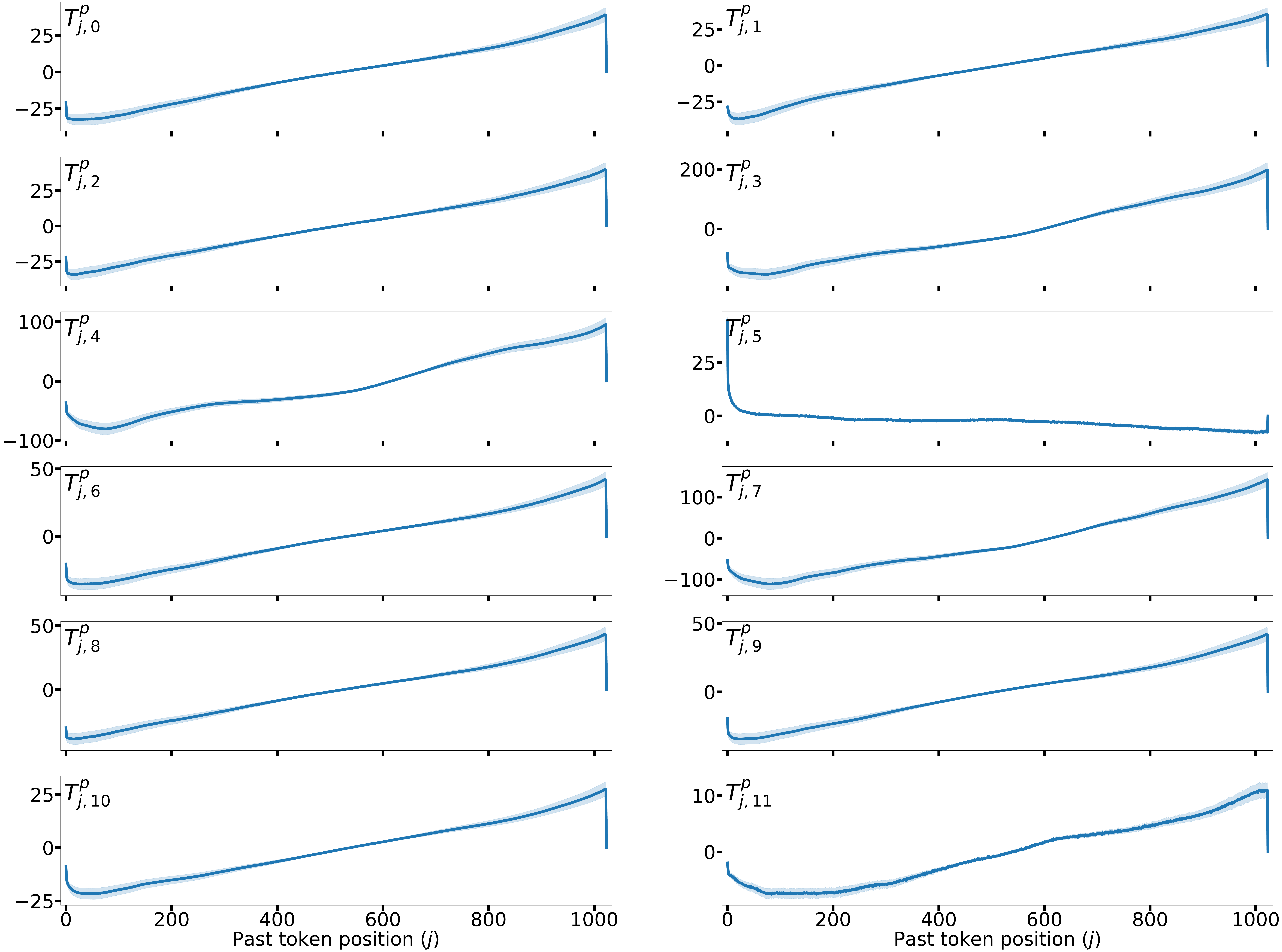}
    \caption{\term{p} for all heads.}
    \label{fig:self-pos-score-all}
\end{figure}

\begin{figure}
    \centering
    \includegraphics[width=\linewidth]{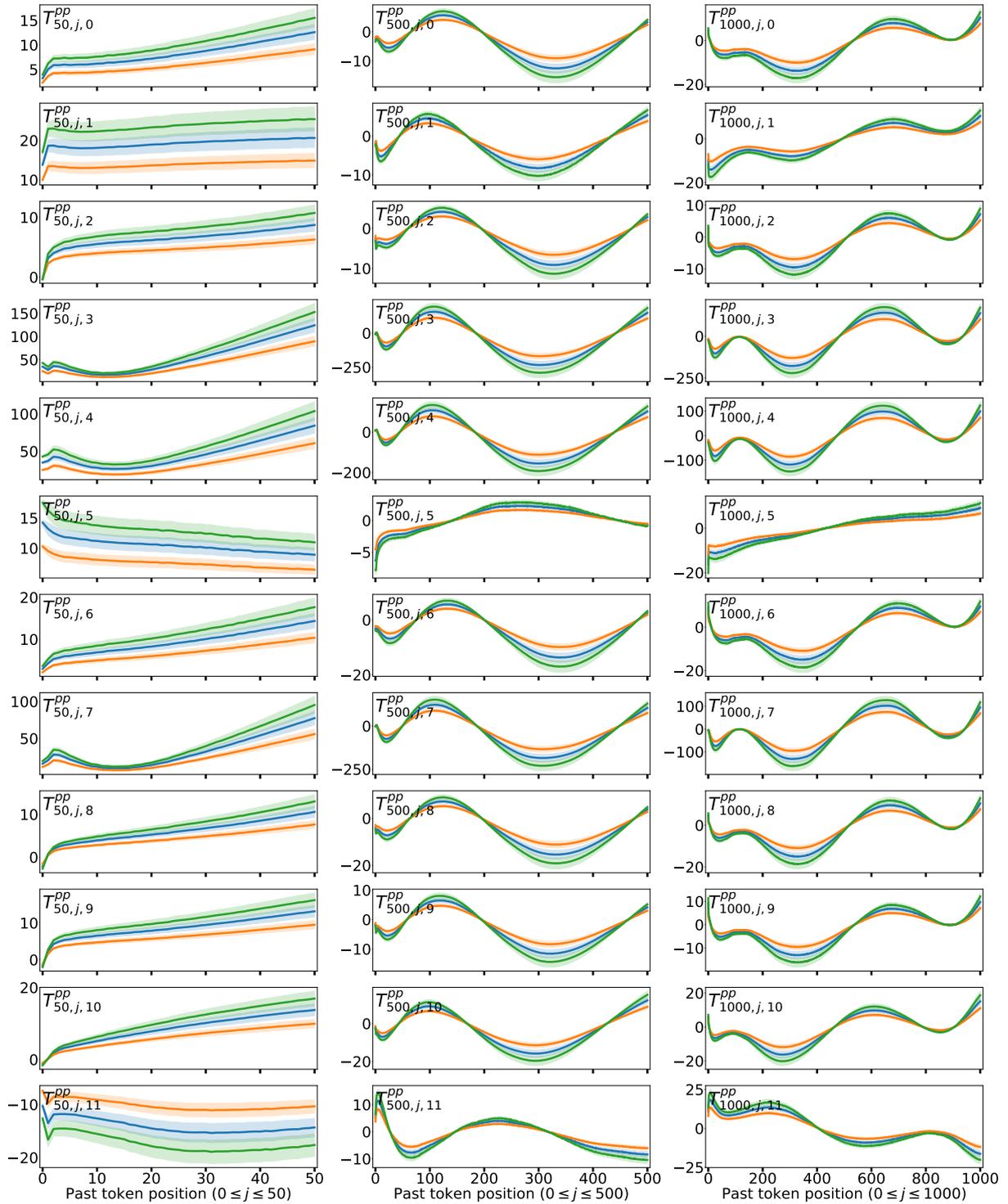}
    \caption{\term{pp} for all heads, for $i\in \{50, 500, 1000\}$.}
    \label{fig:compare-pos-score-all}
\end{figure}

\begin{figure}
    \centering
    \includegraphics[width=\linewidth]{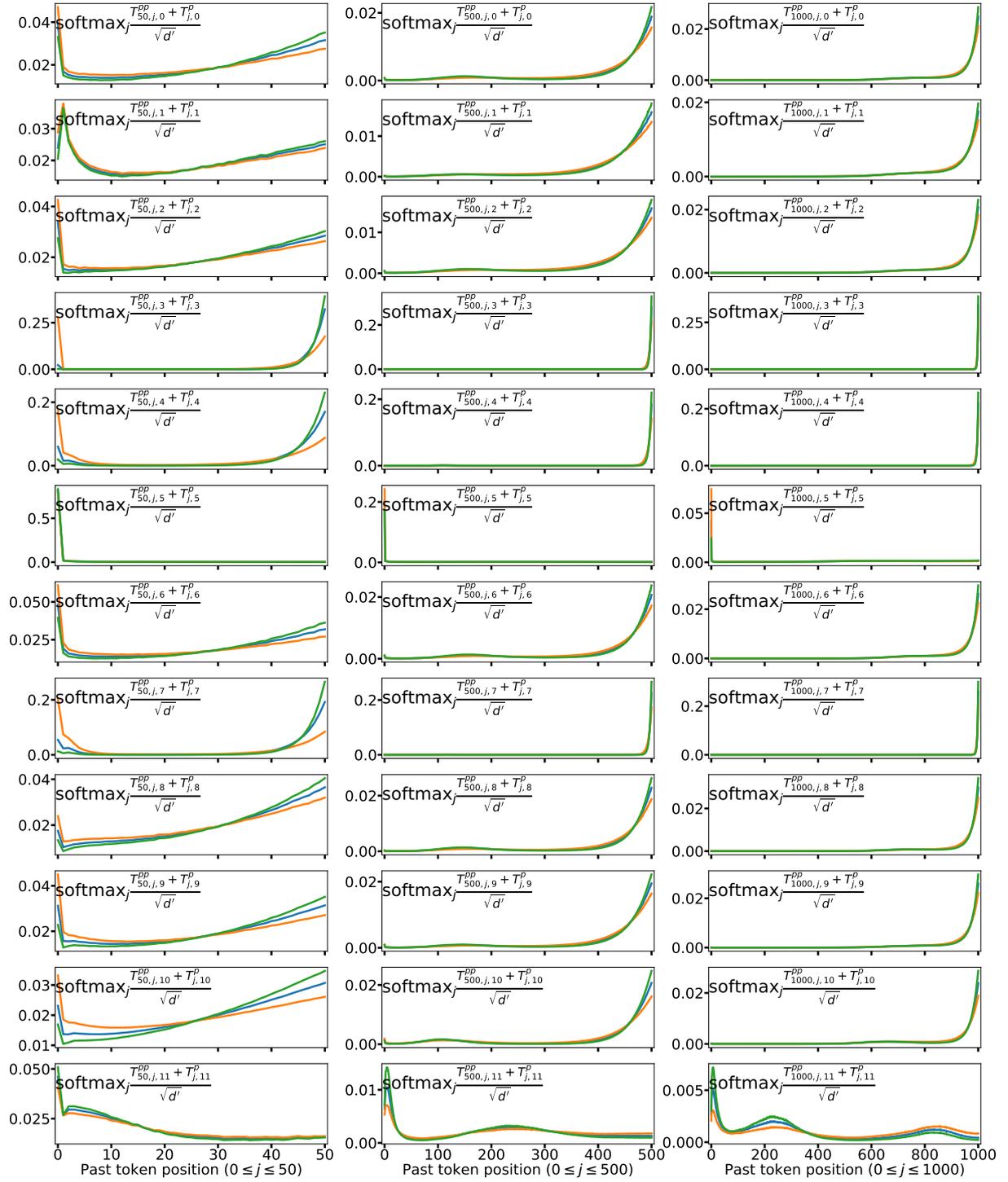}
    \caption{Sum of \term{pp} and \term{p} after softmax with temperature $\sqrt{d'}$ for all heads, for $i\in\{50, 500, 100\}$.}
    \label{fig:both-pos-scores-all}
\end{figure}

\begin{figure}
    \centering
    \includegraphics[width=\linewidth]{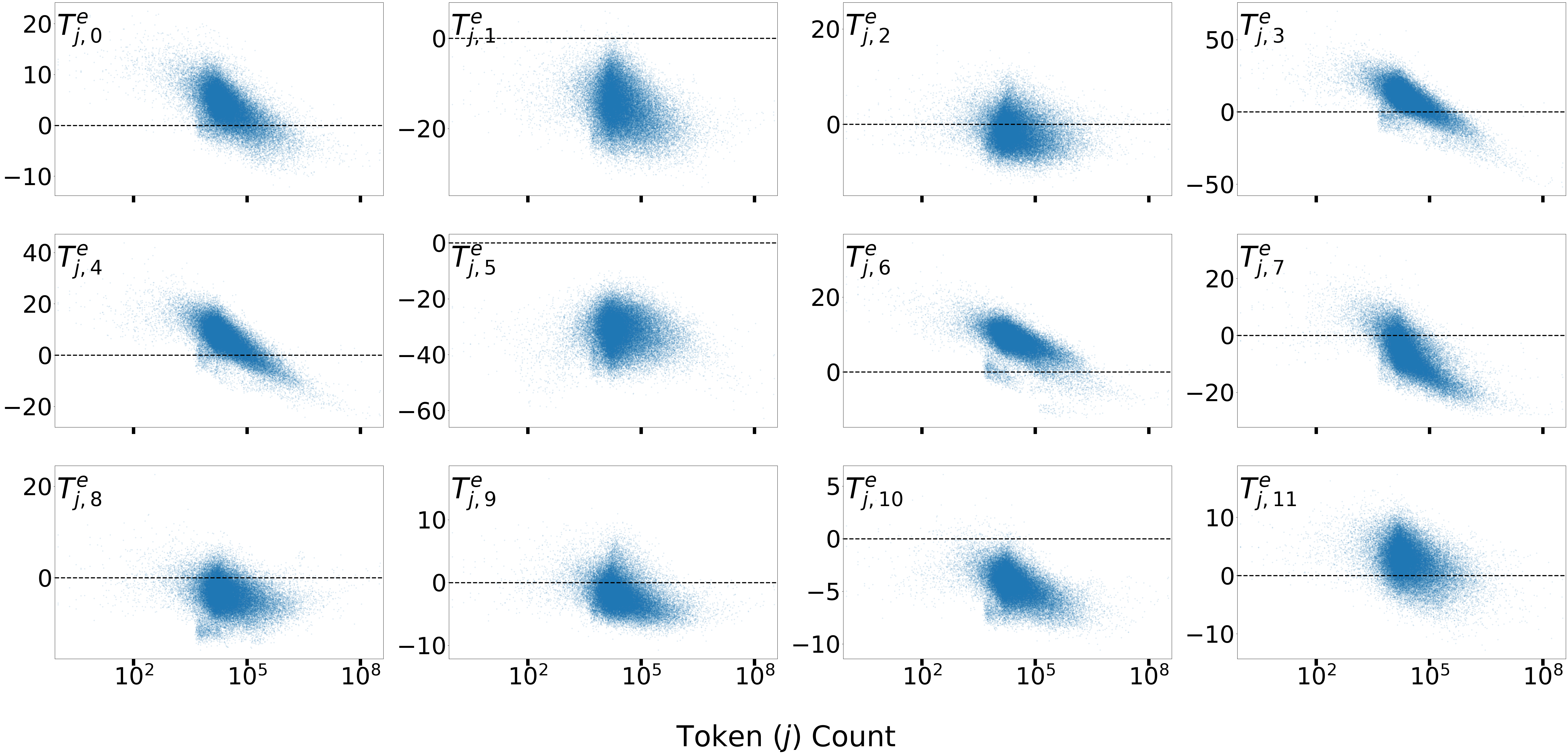}
    \caption{Relation between token frequency and \term{e} for all heads.}
    \label{fig:self-tok-scores-all}
\end{figure}

\begin{figure}
    \centering
    \includegraphics[width=\linewidth]{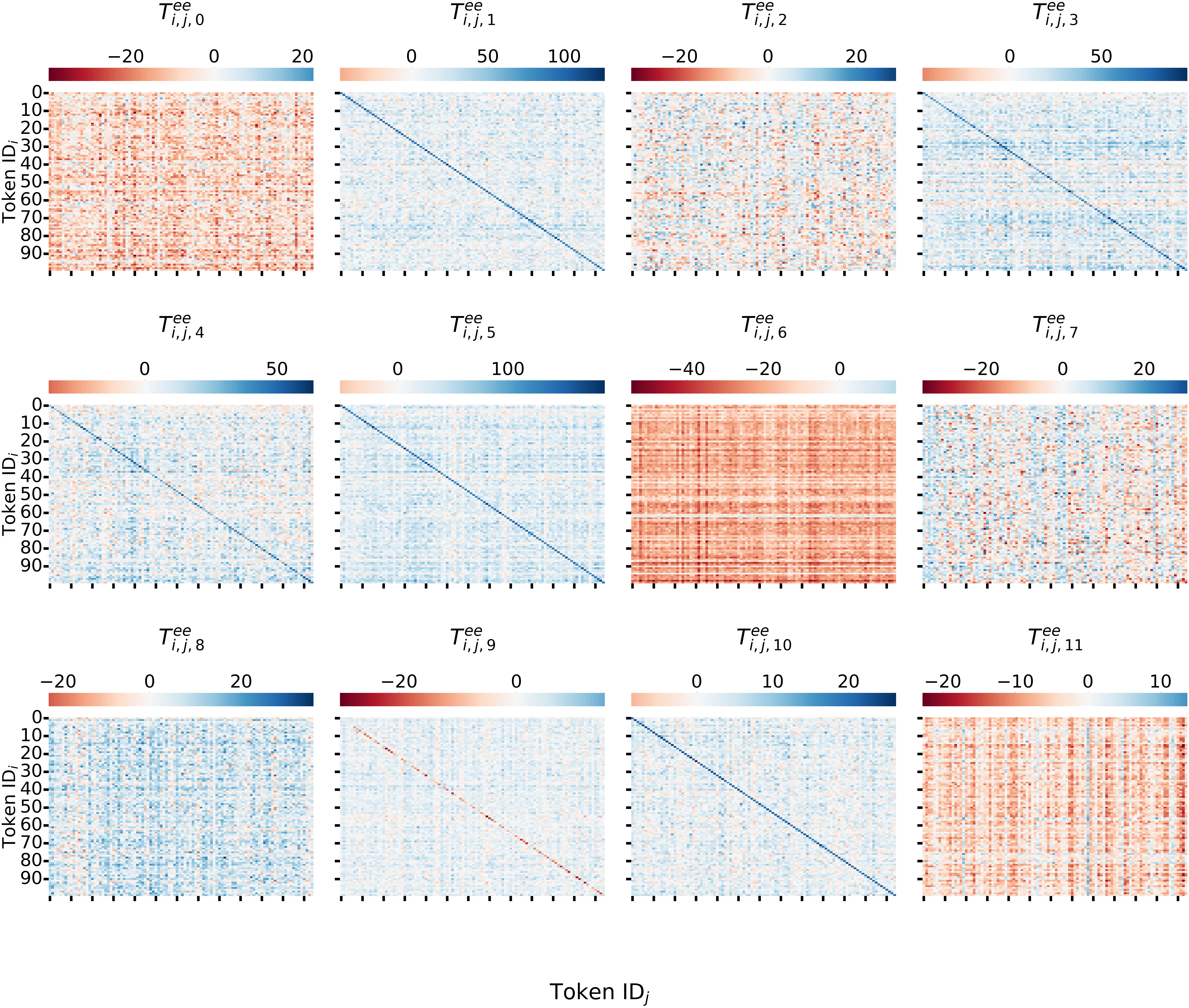}
    \caption{Heatmap of \term{ee} with and without LayerNorm for all heads. Tokens are randomly sampled from the vocabulary for visualization.}
    \label{fig:compare-tok-scores-all}
\end{figure}

\begin{figure}
    \centering
    \includegraphics[width=\linewidth]{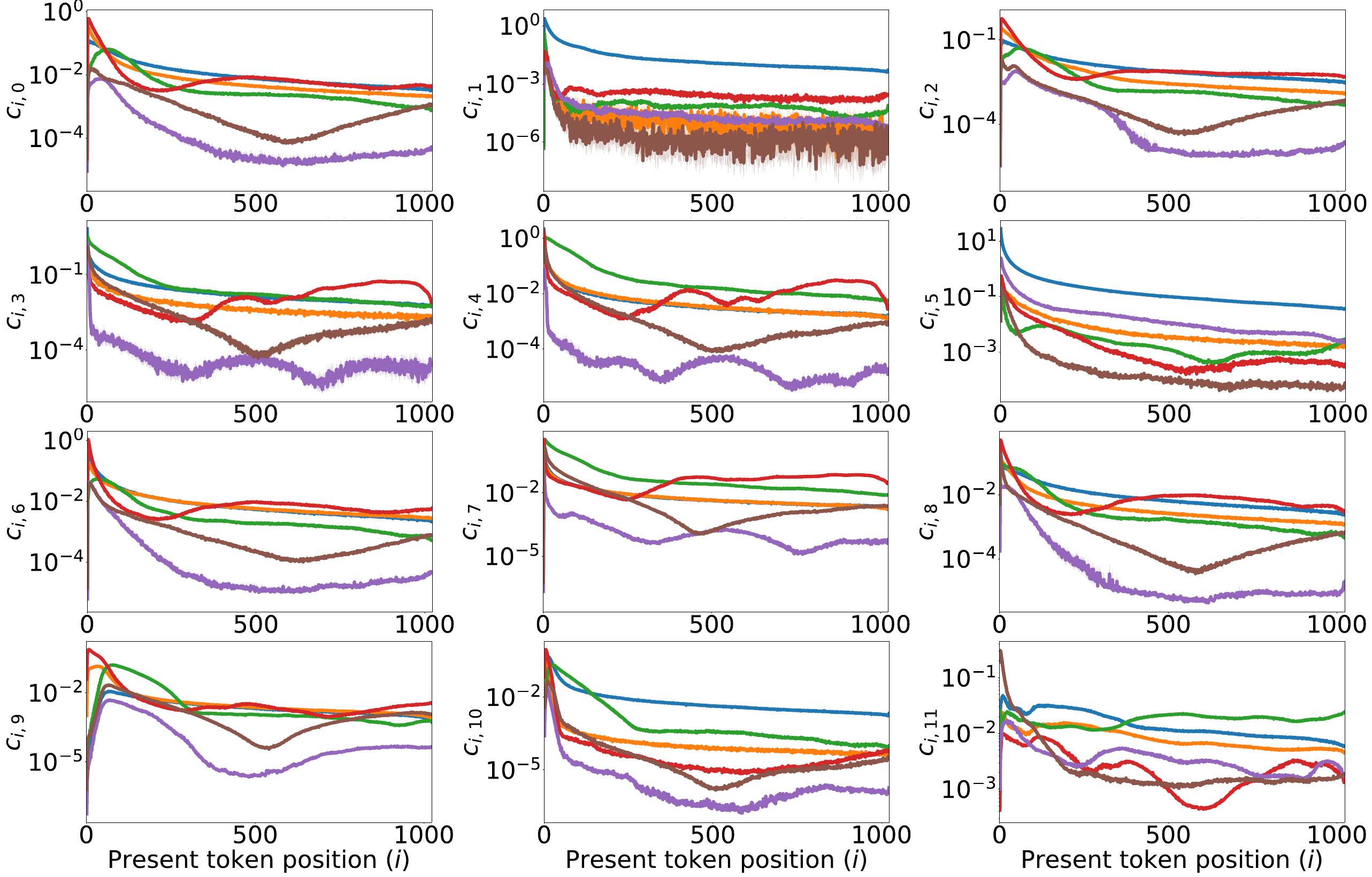}
    \caption{Contribution of the 6 terms in \cref{eq:attn_s_6terms} for each current token position $i$ for all heads.}
    \label{fig:term-contribution-all}
\end{figure}

We show visualizations in \cref{fig:detokenization-token}-\textbf{\textcolor{pptblue}{B}}, \cref{fig:detokenization-position}-\textbf{\textcolor{pptblue}{A},\textcolor{pptblue}{B},\textcolor{pptblue}{D}} and \cref{fig:term-contribution} for all heads in \cref{fig:self-pos-score-all,fig:compare-pos-score-all,fig:both-pos-scores-all,fig:self-tok-scores-all,fig:compare-tok-scores-all,fig:term-contribution-all}.

\newpage

{
    \begin{table}
        \small
        \centering
        \begin{tabular}{clll}
            \toprule
            head & query token & key token (rank) & Resulting sequence\\
            \midrule
            3 & yo & Tok (1) & Tokyo \\
            3 & yo & \_Tok (2) & \_Tokyo \\
            4 & yo & Tok (6) & Tokyo \\
            7 & yo & Tok (1) & Tokyo \\
            4 & \_Korea & \_North (1) & North\_Korea \\
            7 & \_Korea & \_North (1) & \_North\_Korea \\
            7 & \_Korea & North (2) & North\_Korea \\
            7 & \_Korea & \_South (3) & \_South\_Korea \\
            7 & \_Korea & South (4) & \_South\_Korea \\
            1 & \_Obama & \_Barack (3) & \_Barack\_Obama \\
            1 & \_Obama & President (8) & \_President\_Obama \\
            4 & \_Obama & \_Barack (3) & \_Barack\_Obama \\
            5 & \_Obama & \_Barack (5) & \_Barack\_Obama \\
            7 & \_Obama & \_President (2) & \_President\_Obama \\
            7 & \_Obama & \_Michelle (3) & \_Michelle\_Obama \\
            7 & \_Einstein & \_Albert (1) & \_Albert\_Einstein \\
            7 & \_Einstein & Albert (2) & Albert\_Einstein \\
            7 & \_Jackson & \_Michael (1) & \_Michael\_Jackson\\
            7 & \_Jackson & \_Peter (2) & \_Peter\_Jackson\\
            7 & \_Jackson & Michael (3) & Michael\_Jackson\\
            7 & \_Jackson & \_Jesse (4) & \_Jesse\_Jackson\\
            7 & \_Jackson & Peter (5) & Peter\_Jackson\\
            7 & \_Jackson & \_Janet (6) & \_Janet\_Jackson\\
            7 & \_chloride & \_aluminum (1) & \_aluminum\_chloride \\
            7 & \_chloride & \_copper (3) & \_copper\_chloride \\
            7 & \_chloride & \_vinyl (6) & \_vinyl\_chloride \\
            7 & \_chloride & \_sodium (7) & \_sodium\_chloride \\
            7 & \_chloride & \_platinum (8) & \_platinum\_chloride \\
            10 & \_chloride & \_potassium (2) & \_potassium\_chloride \\
            10 & \_chloride & \_sodium (3) & \_sodium\_chloride \\
            3 & \_century & \_19 (1) & \_19\_century \\
            3 & \_century & \_nineteenth (7) & \_nineteenth\_century \\
            7 & \_century & \_21 (1) & \_21\_century \\
            7 & \_century & \_twentieth (6) & \_nineteenth\_century \\
            \bottomrule
        \end{tabular}
        \caption{Exerpt of analysis on \term{ee} which support detokenization.}
        \label{tab:detokenization-all}
    \end{table}
}

